\documentclass[manuscript,screen,nonacm]{acmart}

\def\ShowAuthNotes{1}

\ifnum\ShowAuthNotes=1
	\newcommand{\sai}[1]{\textcolor{blue}{[{\bf Sai:} {#1}]}}
	\newcommand{\osonde}[1]{\textcolor{blue}{[{\bf Osonde:} {#1}]}}
        \newcommand{\rynote}[1]{\textcolor{red}{[{\bf Ryan:} {#1}]}}
        \newcommand{\miao}[1]{\textcolor{blue}{[{\bf Miao:} {#1}]}}
\else
	\newcommand{\sai}[1]{}
	\newcommand{\osonde}[1]{}
        \newcommand{\rynote}[1]{}
        \newcommand{\miao}[1]{}
\fi

\usepackage{amsmath}
\usepackage{amsfonts}

\usepackage{graphicx, xspace}
\usepackage{tabularx}

\newcommand{\sk}{\mathsf{sk}}
\newcommand{\pk}{\mathsf{pk}}
\newcommand{\ct}{\mathsf{ct}}
\newcommand{\M}{\mathcal{M}}
\newcommand{\G}{\mathcal{G}}
\newcommand{\secr}{\lambda}
\newcommand{\RO}{\mathsf{RO}}
\newcommand{\ahesk}{\mathsf{sk_{HE}}}
\newcommand{\ahepk}{\mathsf{pk_{HE}}}
\newcommand{\symsk}{\mathsf{sk_{Sym}}}
\newcommand{\rand}{\mathsf{rand}}
\newcommand{\cS}{\mathcal{S}}
\newcommand{\bT}{\mathsf{T}}

\newcommand{\ahe}{\mathsf{HE}}
\newcommand{\ahekeygen}{\mathsf{HE.KeyGen}}
\newcommand{\aheenc}{\mathsf{HE.Enc}}
\newcommand{\aheadd}{\mathsf{HE.Add}}
\newcommand{\ahedec}{\mathsf{HE.Dec}}

\newcommand{\com}{\mathsf{Com}}
\newcommand{\comkeygen}{\mathsf{Com.KeyGen}}
\newcommand{\comenc}{\mathsf{Com.Enc}}
\newcommand{\comdec}{\mathsf{Com.Dec}}

\newcommand{\symkeygen}{\mathsf{Sym.KeyGen}}
\newcommand{\symenc}{\mathsf{Sym.Enc}}
\newcommand{\symdec}{\mathsf{Sym.Dec}}

\newcommand{\namedref}[2]{\hyperref[#2]{#1~\ref*{#2}}\xspace}
\newcommand{\sectionref}[1]{\namedref{Section}{sec:#1}}
\newcommand{\figureref}[1]{\namedref{Figure}{fig:#1}}
\newcommand{\memid}{\mathsf{mem_{id}}}
\newcommand{\out}{\mathsf{output}}
\newcommand{\racep}{\vec{\mathsf{race_p}}}
\newcommand{\values}{\vec{\mathsf{values}}}

\acmYear{2026}

\begin{document}

\title{Productionized Fairness Measurement Under Privacy Constraints}

\author{Osonde A. Osoba}
\orcid{0000-0002-5232-2779}
\email{oosoba@linkedin.com}
\affiliation{%
  \institution{LinkedIn}
  \state{California}
  \country{USA}
}

\author{Yuzi He}
\email{yuzhe@linkedin.com}
\affiliation{%
  \institution{LinkedIn}
  \state{California}
  \country{USA}
}

\author{Saikrishna Badrinarayanan}
\email{sbadrina@linkedin.com}
\affiliation{%
  \institution{LinkedIn}
  \state{California}
  \country{USA}
}

\author{Varun Mithal}
\affiliation{%
  \institution{LinkedIn}
  \state{California}
  \country{USA}
}

\author{Sakshi Jain}
\affiliation{%
  \institution{LinkedIn}
  \state{California}
  \country{USA}
}

\author{Natesh S. Pillai}
\authornote{Work done while at LinkedIn.}
\affiliation{%
  \institution{LinkedIn}
  \state{California}
  \country{USA}
}

\renewcommand{\shortauthors}{Osoba et al.}

\begin{abstract}
Fairness measurements in the form of disaggregated evaluations often rely on demographic signals that are legally constrained or culturally sensitive.
Race and ethnicity signals are among the more difficult signals to curate and use for this task.
This paper presents Privacy-Preserving Probabilistic Race/Ethnicity Estimation (PPRE) as a method for enabling fairness measurements with respect to race/ethnicity for U.S.\ LinkedIn members in a privacy-preserving manner.
PPRE applies privacy technologies (specifically: secure two-party computation, differential privacy, and additive homomorphic encryption) on top of two race/ethnicity demographic signal sources (the Bayesian Improved Surname Geocoding estimator and a sparse golden survey set of self-reported demographics) to power a fairness measurement solution with respect to US-based race/ethnicity demographics.
We detail its privacy guarantees and demonstrate its application on candidate- and viewer-side fairness measurements.
We close with a transferable framework for institutions seeking to implement similar privacy-preserving measurement infrastructure.
\end{abstract}


\maketitle

\section{Introduction}

Disaggregated systems evaluation is a standard meta-approach for detecting unfair bias~\cite{barocas2021} in AI systems: the evaluator compares and flags any measured significant system performance disparities across demographic groups.
Such evaluations typically require member-level demographic data such as race/ethnicity, gender, or location.
Race/ethnicity presents a difficult case since that demographic signal can be legally sensitive, culturally charged, and often simply unavailable.
Many technology platforms do not collect race/ethnicity data, or hold it only for a small fraction of their user base.
This creates a concrete operational gap: institutions committed to monitoring AI systems for fair treatment across protected demographic groups (including race and ethnicity) cannot fulfill that commitment without first solving a demographic data problem.

Questions involving the concept of race and/or ethnicity can be complex and contested in the United States.
It is worth addressing since policy concerns in the United States often implicate questions of fairness and justice with respect to the admittedly socially-constructed concept of race/ethnicity. 
Rieke et al.~\cite{rieke2022imperfect} expresses this tension as such: ``despite the [socially-]constructed, artificial nature of race, race-based inequality has shaped society and institutions in ways that are undeniably real and deserving of ongoing attention.''
Race/ethnicity group information has specific value for understanding wrongful biases in decision systems including in AI systems.
But that demographic information also carries significant risks of misuse.
These risks are concrete: re-identification of individuals from ostensibly aggregate results, function creep that repurposes a fairness dataset toward discriminatory or otherwise unintended ends, regulatory exposure under anti-discrimination law, and the corrosion of user trust that attends any collection of sensitive attributes.
Each of these harms falls hardest on the very people the measurement is meant to protect, and each grows with the amount of sensitive data assembled and the length of time it is retained.
And yet forgoing measurement with respect to race/ethnicity entirely would leave important biases in AI systems undetectable and unaddressed.
The central design question is therefore this: \emph{how can we enable useful race/ethnicity fairness measurement while limiting the privacy and misuse risks of holding such sensitive demographic information?}

The Privacy-Preserving Probabilistic Race/Ethnicity Estimation (PPRE) approach answers this question by combining probabilistic race/ethnicity estimation with privacy-preservation techniques.
It enables useful fairness measurements without exposing sensitive demographic information.
PPRE is deployed at production-scale to measure AI systems for race/ethnicity fairness while preserving the privacy of member data.

\subsection{LinkedIn-specific Version of the Problem}
\label{sec:problem}

LinkedIn leverages AI at scale for use cases such as job recommendations, content ranking, and recruiter search.
LinkedIn's Responsible AI Principles~\cite{LawitXu_2023} include a commitment to ``Promote Fairness and Inclusion: i.e. ensuring that the use of AI benefits all members fairly, without causing or amplifying unfair bias.''
Fulfilling this commitment requires measuring AI system performance along key demographic dimensions, including race/ethnicity.
At LinkedIn, race/ethnicity attributes are available for approximately 6\% of U.S.\ members through a voluntary Self-ID survey\footnote{\url{https://members.linkedin.com/equal-access}}.
This sample is too sparse for general fairness measurements. 
Furthermore, it is a self-selected sample, introducing \emph{selection effects} that prevents generalization to the broader population.
To address this lack of demographic data, we examined and rejected several alternative methods.
One alternate path, \emph{debiasing based on propensity-scores}, applied causal and quasi-experimental methods to try to remove selection bias from AI fairness measurements based on Self-ID samples.
Debiased estimates showed the correct direction of bias but magnitude accuracy proved difficult to calibrate. 
Getting the measurement magnitude right is pivotal for downstream mitigation efforts.
Furthermore, this measurement approach would require recalibrating propensity score models for each AI system under testing. This is not a scalable measurement strategy especially as the portfolio of candidate AI system explodes both in size and variety.
We likewise rejected early system designs that persisted demographic estimates to a common
location after applying both local DP to Self-ID records and encryption-at-rest.
This path was rejected because it required persisting sensitive data or performing identifiable plaintext analysis.

\subsection{Design Principles \& Evolution}
\label{sec:principles}

The final design (PPRE) emerged after several iterations, informed by consultations with internal stakeholders and external experts~\cite{hai-wrkshp2023}.
These explorations crystallized a key requirement: all member-level demographic data must be \emph{ephemeral}, generated on-demand, encrypted immediately, and deleted after aggregation.
Four core privacy principles emerged:
\begin{enumerate}
    \item \emph{Data minimization}: use the least amount of personal data needed, respecting LinkedIn's privacy commitments and the sensitivities members may have around race/ethnicity information.
    \item \emph{No individual race/ethnicity assignment}: avoid assigning a single race or ethnicity category to any member. Only probabilistic estimates are computed, and they are ephemeral: computed on-demand, never stored on disk, encrypted by design, and aggregated.
    \item \emph{Strong security}: implement privacy and security protections to prevent unauthorized access to the measurement workflow, including efforts to prevent reassociation of estimate calculations to identifiable persons.
    \item \emph{No persistent sensitive associations}: avoid solutions that create a potentially persistent association of members with demographic labels, especially since estimated demographic information can be incorrect and/or imprecise. This ruled out approaches that would store a table of deliberately falsified race data (e.g., applying local DP to Self-ID records and persisting the result). 
\end{enumerate}

The design that satisfied these principles is a \emph{secure two-party computation protocol}.
Two parties, each holding a different data component, jointly compute fairness measurements without observing each other's plaintext inputs.
Demographic data is generated ephemerally and immediately encrypted; only aggregate statistics are decrypted.
The protocol composes commutative encryption, additive homomorphic encryption, and differential privacy to balance utility, computation cost, communication cost, and privacy.

\subsection{The PPRE Method}
\label{sec:approach}

PPRE rests on three conceptual pillars.
\emph{Demographic signal processing} that generates and interprets probabilistic race/ethnicity estimates.
\emph{Privacy primitives and their extensions} to provide the cryptographic and statistical toolkit.
And a \emph{secure computation protocol} that composes them into an end-to-end system.

The first pillar is the \emph{demographic signal processing} that spans the full information pipeline from raw inputs through to the Bayesian estimation process all the way to generation of interpretable fairness measurements.
We rely on the BISG method~\cite{elliott2009, imai2022} to power this processing flow. 
BISG is a U.S. Census-normed Bayesian model that produces \emph{probabilistic} estimates of race/ethnicity from surname $S$ and location $G$.
It computes the posterior $\Pr(R | S, G)$ over six categories~\cite{omb_1997} using 2010 U.S.\ Census frequency tables~\cite{census_surnames_2016, census_summary_2012}.
Critically, BISG outputs a probability distribution rather than a categorical assignment.
This satisfies the no-individual-assignment principle while enabling aggregate measurement.
The Self-ID survey provides a sparse calibration signal: voluntary self-reports from approximately 6\% of U.S.\ members.
These are privatized and combined with BISG estimates before use (\sectionref{pets}, \sectionref{ppre-system}).
Furthermore, since PPRE operates on probability distributions rather than categorical labels, standard disparity estimators must be generalized.
Following examples like Chen et al.~\cite{chen2019} and Elzayn et al.~\cite{elzayn2023}, we replace hard indicators $\mathbb{I}(R_i = j)$ with soft indicators $\Pr(R_i = j)$.
This gives weighted-average estimators compatible with probabilistic demographics.
\sectionref{audit-func} develops the probabilistic disparity estimator framework and applies it to three deployed fairness metrics.

The second pillar is the \emph{privacy primitives and extensions} thereon that provide the cryptographic and statistical tools, together with practical extensions that broaden their reach.
Three core primitives underpin the protocol.
\emph{Commutative encryption} (Pohlig-Hellman on Curve25519~\cite{Ber06}) enables private set intersection via order-independent doubly-encrypted identifiers.
\emph{Additive homomorphic encryption} (Paillier~\cite{Paillier99}, 2048-bit modulus) allows computation on ciphertexts, supporting the linear aggregations that probabilistic disparity estimators require.
\emph{Local differential privacy} (randomized response~\cite{Warner65}) protects Self-ID records before they enter the protocol, preventing re-identification across measurement runs.

Estimators that are linear functions of probabilities and test values are directly computable under Additive homomorphic encryption (AHE).
The set of operators required for production fairness metrics extends beyond just linear functions.
Fairness measurement depends heavily on nonlinear operators like the division, value comparison, floating-point arithmetic, and hypothesis testing operations that production fairness metrics demand.
AHE does not natively support these non-linear operations so we had to develop adaptations to the native AHE protocol to enable these operations.
\sectionref{he-adapt} discusses protocol adaptations that enable computation for disparity metrics that are not strictly linear.
Our extensions to the standard AHE toolkit comprise five adaptation patterns: ratio estimation via separate numerator/denominator aggregation, fixed-point encoding, pre-encryption comparison, signed-number encoding, and bootstrap hypothesis testing (\sectionref{he-adapt}).
Together, these extensions enable production metrics including the Listwise Outcome Test (LOT) and Minimum Quality of Service -- NDCG (MQOS-NDCG).

The third pillar is the \emph{protocol composition} that combines the demographic signal processing and privacy tools into a secure two-party computation protocol.
The protocol operates between \emph{$P_1$ (the Tester)}, the Responsible AI team holding privatized demographics, and \emph{$P_2$ (the Test Client)}, the product team holding member IDs, model scores, and outcomes.
$P_1$ prepares the demographic dataset by applying local DP to Self-ID records, computing BISG estimates, combining them, and applying probabilistic clipping to prevent near-certain assignments.
Both parties then execute commutative encryption on their join-keys (required for PSI) and other encryption-related operations for their respective value columns.
The parties execute a PSI-based encrypted join followed by homomorphic aggregation, facilitated by a sequence of encrypted table exchanges.
$P_1$ computes weighted sums over $P_2$'s encrypted test values without decryption.
$P_2$ then decrypts the aggregated sums to obtain the final plain-text disparity statistics.
Only aggregated disparity statistics are ever decrypted.

The system's guarantees are the combined effect of these three pillars:
No individual is assigned a race/ethnicity.
The two parties exchange only encrypted data, with member identifiers doubly encrypted and discarded before computation.
Neither party learns the complete tuple of identity, demographics, and model features.
Local DP with probabilistic clipping prevents re-identification of Self-ID respondents across runs, and all member-level data is deleted after processing.
Further governance controls, including legal approval requirements, minimum population thresholds, and prohibitions on specific measurements, help defend against other concerns like differencing attacks.

The PPRE system is deployed in production for race/ethnicity fairness measurements of AI ranking systems, enabling both candidate- and viewer-side tests on two-sided recommendation markets.
This production experience provides an existence proof that privatized multi-party fairness measurement is feasible at industrial scale.
It also offers practical context for cross-organizational secure computation architectures frequently proposed in the responsible AI literature.

\subsection{Contributions \& Related Work}
\label{sec:contributions}

PPRE hybridizes methods from multiple fields into a single production system.
This paper contributes along each of the three pillars:
\begin{enumerate}
    \item \emph{Demographic signal processing.}
    We develop probabilistic disparity estimators that generalize standard fairness metrics to operate on BISG probability distributions rather than categorical labels.
    BISG originates in healthcare disparities research~\cite{elliott2009} and has since been applied in fair lending~\cite{voicu2018, cfpb2014}, ad fairness~\cite{meta-ads2023}, redistricting~\cite{deluca-curiel2023}, and imputation quality assessments~\cite{imai2022, rieke2022imperfect}.
    Known limitations include age-related degradation~\cite{rieke2022imperfect}, downstream estimator bias~\cite{chen2019, kallus2022}, and narrow categorization~\cite{pai-demog2021} (see \sectionref{bisg}).
    Extending work by Chen et al.~\cite{chen2019} and Kallus et al.~\cite{kallus2022} on fairness with unobserved or probabilistic protected attributes, we develop estimators for previously-unpublished candidate- and viewer-side fairness tests on two-sided recommendation markets.

    \item \emph{Privacy primitives and extensions.}
    We describe a catalog of practical HE adaptations (\sectionref{he-adapt}) that extend additive HE to support the non-linear operations fairness metrics require.
    The underlying primitives have extensive foundations~\cite{Yao82, Meadows86, HFH99, Paillier99, PSWW18, IKNPRSSSY19, BKMS20, Warner65, EvfimievskiGeSr03, KasiviswanathanLeNiRaSm11, DworkMcNiSm06}.
    Our contribution is composing them into an effective protocol and extending them with sampling-based uncertainty quantification to avoid non-linear operations on encrypted data.

    \item \emph{Protocol composition.}
    We detail a custom 2PC protocol composing commutative encryption, additive HE, and local DP.
    We report on its production deployment including fairness measurements of two AI ranking systems.
\end{enumerate}

Together, these contributions constitute PPRE: a complete, deployed system for privacy-preserving race/ethnicity fairness measurement with explicit design principles and rationale.
The design patterns, protocol architecture, and engineering solutions documented here form a transferable framework for institutions seeking similar privacy-preserving measurement infrastructure.

\paragraph{Relationship to prior work.}
This paper builds on an earlier report~\cite{ppre-arxiv2024} that introduced the PPRE architecture, defined the cryptographic primitives, and validated the estimator framework in plaintext.
That earlier work described the protocol without the HE adaptations that production metrics require (\sectionref{he-adapt}).
Its demonstrations explicitly ``forego the homomorphic encryption part of PPRE to simplify numerical comparisons.''
The present paper bridges these gaps.
We develop the HE adaptations catalog that makes encrypted fairness computation practical, specify the concrete metric algorithms (ERO, LOT, MQOS-NDCG) as deployed instantiations of the general estimator framework, validate the full encrypted pipeline end-to-end, and report production fairness measurements of two AI ranking systems.

\paragraph{Industry precedents.}
Two industry systems address related problems.
Airbnb's Project Lighthouse~\cite{LighthouseAirbnb_2020} measures booking acceptance disparities using an external computation facilitator, syntactic disclosure control ($k$-anonymity, $p$-sensitivity), and third-party demographic estimation.
PPRE differs in three ways: it operates entirely within one organization, uses differential privacy instead of syntactic anonymity, and uses BISG.
Meta's Variance Reduction System (VRS)~\cite{meta-ads2023, meta-ppre} also uses BISG and DP for ad delivery fairness measurement, but applies DP to aggregated output and relies on external facilitators.
PPRE applies DP at the input stage, to Self-ID records before they enter the protocol.
This provides a stronger guarantee: each individual's data is protected before combination or computation rather than after.

\paragraph{Paper structure.}
\label{sec:structure}
The remainder of the paper presents each tool before the application that needs it.
\emph{Privacy primitives and extensions}: \sectionref{pets} defines the core cryptographic and statistical tools (commutative encryption, homomorphic encryption, differential privacy, and key management); \sectionref{he-adapt} catalogs general-purpose HE adaptations that extend their reach to practical fairness metrics.
\emph{Demographic signal processing}: \sectionref{f-audits} presents the BISG estimation method; \sectionref{metrics} develops probabilistic disparity estimators and the HE computation strategy for each deployed fairness metric.
\emph{Protocol composition}: \sectionref{ppre-system} presents the full two-party computation protocol that composes these elements.
\sectionref{impl} reports on the production deployment, including validation analyses and fairness measurements of two AI ranking systems.
\sectionref{disc} discusses operational learnings, limitations, and future work.
\sectionref{conc} summarizes our contributions.

\section{Privacy-Enhancing Technologies}\label{sec:pets}
This section presents the cryptographic and statistical toolkit that PPRE draws on.
Three core primitives form its foundation: commutative encryption, additive homomorphic encryption, and local differential privacy.
The key management regime governing all three is also defined here.
\sectionref{he-adapt} extends this toolkit with adaptation patterns that broaden what additive HE can compute.
\sectionref{ppre-system} composes the full set into a two-party protocol.
Formal definitions of all primitives appear in Appendix~\ref{sec:appendix-primitives}; we summarize the essential properties here and refer the reader there for complete specifications.
Throughout this paper, $\secr$ denotes the computational security parameter.

\subsection{Secure Two-Party Computation}
Secure two-party computation (2PC) enables two parties with private datasets to jointly compute a function on their combined inputs without revealing anything beyond the function's output~\cite{Yao82, Lindell21}.
PPRE's protocol relies on two cryptographic primitives: commutative encryption and additive homomorphic encryption.

\paragraph{Commutative encryption.}
PPRE uses commutative encryption to enable private set intersection: the two parties identify shared members without revealing their respective member lists.
The key property is that encrypting a message under two keys produces the same ciphertext regardless of the order the keys are applied:
$\comenc_{\sk_1}(\comenc_{\sk_2}(m)) = \comenc_{\sk_2}(\comenc_{\sk_1}(m))$.
Combined with a fresh random oracle $\RO$ instance per protocol execution, this serves as the building block for PSI protocols~\cite{Meadows86, HFH99}.
We instantiate commutative encryption as the Pohlig-Hellman cipher over Curve25519~\cite{Ber06}.

\paragraph{Homomorphic encryption.}
PPRE's fairness metrics are weighted averages of encrypted test values.
Additive homomorphic encryption~\cite{Paillier99} supports exactly this: additions and scalar multiplications of encrypted messages without the secret key.
Given ciphertexts $\aheenc(m_1)$ and $\aheenc(m_2)$, one can compute $\aheenc(m_1 + m_2)$ via homomorphic addition and $\aheenc(c \cdot m)$ via scalar multiplication for any constant $c$.
These two operations suffice for the linear aggregations that PPRE's disparity estimators require.
But production fairness metrics also demand division, comparison, and hypothesis testing. \sectionref{he-adapt} addresses these gaps.

\subsection{Differential Privacy}
The third primitive addresses a different threat.
Commutative encryption and HE protect data in transit and during computation, but they do not protect against inference from repeated protocol executions.
\emph{Differential Privacy} (DP)~\cite{DworkMcNiSm06} addresses this by adding calibrated randomness so that an adversary cannot determine with certainty whether any individual's data was used in a computation.
This protects Self-ID respondents from inference attacks that could arise from repeated protocol runs, even if the underlying cryptographic primitives remain secure.

We use the \emph{local model}, which applies randomness to each individual's data before aggregation, providing stronger privacy at the cost of lower utility compared to the global model.
An algorithm $\mathcal{M}: \mathcal{X} \rightarrow \mathcal{Y}$ is $\epsilon$-locally DP~\cite{EvfimievskiGeSr03, Warner65, KasiviswanathanLeNiRaSm11} if, for all inputs $x, x' \in \mathcal{X}$ and all outcomes $\cS \subseteq \mathcal{Y}$:
$$\mathsf{Pr}[\mathcal{M}(x) \in \cS] \leq e^\epsilon \mathsf{Pr}[\mathcal{M}(x') \in \cS].$$
We apply the traditional randomized response mechanism~\cite{Warner65} to Self-ID records before they enter the protocol.
Each Self-ID record is a one-hot vector over $k=6$ race/ethnicity categories.
Randomized response retains the true value with probability $\frac{e^\epsilon}{e^\epsilon + k - 1}$ and flips it to one of the other $(k-1)$ entries with probability $\frac{1}{e^\epsilon + k - 1}$ each.
This gives a concrete, tunable privacy guarantee for each Self-ID record.

\subsection{Key Management}
\label{sec:key-management}

Cryptographic keys are the most sensitive element in the protocol.
If keys persist beyond a single computation session, or if they are accessible to the wrong party, the protocol's privacy guarantees collapse.
The PPRE protocol enforces strict controls on all cryptographic material:
\begin{itemize}
    \item \textbf{Locally generated.} Each party's computation job generates its own random keys internally; keys are not provisioned or distributed by external entities.
    \item \textbf{Ephemeral.} Keys are used for one computation session only and destroyed at session end. And, for encryption operations for which we never need to decrypt (e.g., the commutative encryption for PSI), we do not even materialize a decryption key.
    \item \textbf{In-memory only.} Keys are held exclusively in the computation job's working memory and are never written to persistent storage.
    \item \textbf{No decryption key for member IDs.} The commutative encryption scheme is used only for equality testing during the join. The join does not require decryption, so no key for recovering plaintext member identifiers is ever generated.
    \item \textbf{Separate per party.} Each party generates its own keys independently.  No key is accessible to the other party.
    \item \textbf{Actively managed.} Keys are deleted during the computation as soon as they are no longer needed (e.g., $P_1$ deletes its symmetric key after the test computation step).
\end{itemize}
These constraints ensure that even a post-hoc compromise of one party's storage reveals no cryptographic material from prior measurement sessions.

\section{Adapting Homomorphic Encryption for Fairness Computation}
\label{sec:he-adapt}

Additive homomorphic encryption (\sectionref{pets}) natively supports only addition and scalar multiplication on ciphertexts.
These operations suffice for linear combinations, but production fairness metrics also require division, comparison, floating-point arithmetic, signed arithmetic, and hypothesis testing.
None of those operations are natively available in ciphertext space.
The question is how to bridge this gap without abandoning the privacy guarantees that HE provides.

This section extends the privacy-primitives toolkit with five general-purpose adaptation patterns that broaden what additive HE can compute.
Each pattern addresses a specific operational limitation and is presented at an abstract level.
\sectionref{metrics} shows how specific fairness metrics compose these patterns within the secure computation protocol.

\begin{table*}[ht!]
\centering
\caption{Catalog of HE adaptation patterns.}
\label{tab:he-catalog}
\begin{tabularx}{\textwidth}{@{}l X X@{}}
\toprule
\textbf{Pattern} & \textbf{Problem Addressed} & \textbf{Strategy} \\
\midrule
Ratio estimation (\ref{sec:ratio-estimation}) & Division not supported in HE & Aggregate numerator and denominator separately; divide after decryption \\
Fixed-point encoding (\ref{sec:fixed-point}) & Paillier operates on integers only & Scale reals by $10^c$; track exponent through operations \\
Pre-encryption comparison (\ref{sec:pre-encryption-comparison}) & Comparison/sorting not supported in HE & Exploit data partition: $P_2$ computes comparisons in plaintext before encrypting \\
Signed-number encoding (\ref{sec:signed-encoding}) & Paillier plaintext space is non-negative & Partition $\mathbb{Z}_n$ into positive, negative, and overflow regions \\
Bootstrap hypothesis testing (\ref{sec:bootstrap}) & Variance estimation requires squaring & Resample and aggregate using ratio estimation; construct confidence intervals from bootstrap distribution \\
\bottomrule
\end{tabularx}
\end{table*}

\subsection{Ratio Estimation via Separate Aggregation}
\label{sec:ratio-estimation}

Every fairness metric in PPRE takes the form of a ratio of group-conditional aggregates:
\begin{equation}
    \mu[j] = \frac{\sum_{i=1}^{n} \mathsf{Pr}(R_i = j) \cdot v_i}{\sum_{i=1}^{n} \mathsf{Pr}(R_i = j)}
    \label{eq:ratio-form}
\end{equation}
where $v_i$ is an encrypted test value held by $P_2$ and $\mathsf{Pr}(R_i = j)$ is a plaintext race/ethnicity probability held by $P_1$.
Division cannot be performed in ciphertext space, so the ratio must be decomposed into separately computable parts:
\begin{enumerate}
    \item \textbf{Column expansion.} $P_2$ augments its encrypted table with separate columns for the numerator signal $v_i$ and a denominator column initialized to $1$.

    \item \textbf{HE aggregation.} After the encrypted join, $P_1$ computes two ciphertext aggregates per group $j$:
    \begin{align*}
        \aheenc(\text{Num}[j]) &= \bigoplus_{i=1}^{n} \mathsf{Pr}(R_i = j) \odot \aheenc(v_i) \\
        \aheenc(\text{Den}[j]) &= \bigoplus_{i=1}^{n} \mathsf{Pr}(R_i = j) \odot \aheenc(1)
    \end{align*}
    where $\oplus$ and $\odot$ denote homomorphic addition and scalar multiplication respectively. Both operations use only the native capabilities of the additive HE scheme.

    \item \textbf{Random masking.} Before transferring the encrypted aggregates to $P_2$ for decryption, $P_1$ multiplies both the numerator and denominator ciphertexts by a fresh random nonzero integer $q$ drawn independently for each group $j$.
    This prevents $P_2$ from learning the raw numerator or denominator values.

    \item \textbf{Plaintext division.} $P_2$ decrypts and divides:
    \begin{equation*}
        \mu[j] = \frac{q \cdot \text{Num}[j]}{q \cdot \text{Den}[j]} = \frac{\text{Num}[j]}{\text{Den}[j]}.
    \end{equation*}
    The random factor $q$ cancels exactly, yielding the correct ratio while revealing neither constituent to $P_2$.
\end{enumerate}

This pattern applies to every ratio-form metric in PPRE, including false positive rates, group-conditional NDCG, and listwise outcome tests.
It is the most fundamental adaptation in the catalog.

\subsection{Fixed-Point Encoding}
\label{sec:fixed-point}

The Paillier cryptosystem operates over the integer space $\mathbb{Z}_n = \{0, 1, \ldots, n-1\}$, but BISG probabilities and many test values are real-valued.
The adaptation scales all real-valued inputs to integers before encryption: each value $x \in \mathbb{R}$ is encoded as $\lfloor x \cdot 10^c + 0.5 \rfloor$, where $c$ is a precision parameter.
The protocol tracks the effective scaling exponent through subsequent operations.
For example, a scalar multiplication by a value scaled by $10^c$ produces a result scaled by $10^{2c}$.
The accumulated scale factor is removed after decryption.

\subsection{Signed-Number Encoding}
\label{sec:signed-encoding}

The Paillier plaintext space $\{0, 1, \ldots, n-1\}$ represents only non-negative integers.
Several PPRE metrics, notably the Listwise Outcome Test (LOT), involve score differences that can be negative.

The adaptation partitions the plaintext space into three regions:
\begin{center}
\begin{tabular}{ll}
    $\{0, \ldots, \lfloor n/3 \rfloor\}$ & Non-negative numbers \\
    $\{\lfloor 2n/3 \rfloor + 1, \ldots, n-1\}$ & Negative numbers (encoded as $x + n$) \\
    $\{\lfloor n/3 \rfloor + 1, \ldots, \lfloor 2n/3 \rfloor\}$ & Overflow detection zone \\
\end{tabular}
\end{center}
A negative value $x < 0$ is encoded as $x + n$, placing it in the upper third of the range.
Homomorphic addition of a positive encoding $a$ and a negative encoding $(b + n)$ gives $(a + b + n) \bmod n = (a + b) + n$.
This result correctly represents $a + b$ when $a + b < 0$, or $a + b$ directly when $a + b \geq 0$.
The middle third acts as an overflow buffer: a decrypted value in this range signals that the accumulated sum has exceeded the representable magnitude.
After decryption, values in the upper third are mapped back to their negative counterparts by subtracting $n$.

\subsection{Pre-Encryption Comparison}
\label{sec:pre-encryption-comparison}

Comparison operations such as thresholding ($v_i > t$), sorting, and ranking cannot be performed on ciphertexts under additive HE.
But they can be restructured to occur in \emph{plaintext before encryption} when one party holds the relevant data.
PPRE's natural data partition satisfies this condition: $P_2$ holds model scores, ground-truth labels, and rankings (the data that requires comparisons), while $P_1$ holds race/ethnicity probabilities (which enter only as linear weights).
$P_2$ therefore pre-sorts by score and pre-computes threshold indicators (e.g., $\mathbb{I}(s_i > t)$) and label-derived quantities (e.g., confusion matrix indicators, IDCG-normalized relevance values) before encryption.
The encrypted indicators enter $P_1$'s weighted aggregation as binary or scalar values, requiring only native additive operations.

\subsection{Bootstrap Hypothesis Testing}
\label{sec:bootstrap}

Standard parametric hypothesis tests require squaring, square roots, and division, none of which additive HE supports.
The question is how to assess statistical significance without these operations.
PPRE addresses this by relying on \emph{bootstrap hypothesis testing} procedures, which construct confidence intervals through \emph{resampling} and aggregation rather than parametric formulas.
The bootstrap's core operations, resampling with replacement and computing sample means, decompose into the ratio-estimation pattern of \sectionref{ratio-estimation}.

The adapted bootstrap proceeds through five steps:
\begin{enumerate}
    \item \textbf{Resampling.} From the final deidentified, encrypted joint table, draw $B$ bootstrap samples (e.g., $B = 1000$) by sampling rows with replacement. Each bootstrap sample has the same size as the original dataset.

    \item \textbf{HE-compatible statistic computation.} For each bootstrap sample $b \in \{1, \ldots, B\}$, compute the test statistic using the ratio-estimation pattern:
    \begin{equation*}
        \aheenc(\text{Num}_b[j]), \quad \aheenc(\text{Den}_b[j]) \quad \text{for each group } j.
    \end{equation*}
    Each pair requires only homomorphic addition and scalar multiplication.

    \item \textbf{Decryption and division.} Transfer all $B \times k$ encrypted pairs to $P_2$. $P_2$ decrypts and computes the plaintext statistics:
    \begin{equation*}
        \hat{\theta}_{b}[j] = \frac{\text{Num}_b[j]}{\text{Den}_b[j]}, \quad b = 1, \ldots, B, \; j = 1, \ldots, k.
    \end{equation*}

    \item \textbf{Confidence interval construction.} Sort the $B$ values $\hat{\theta}_{1}[j], \ldots, \hat{\theta}_{B}[j]$ for each group $j$. The $\alpha/2$ and $(1 - \alpha/2)$ percentiles form a $(1-\alpha)$-level bootstrap confidence interval.

    \item \textbf{Hypothesis test.} If the confidence interval for group $j$ excludes a reference value (e.g., a fairness threshold $\tau$ or the overall population statistic), reject the null hypothesis that group $j$'s metric equals the reference at significance level $\alpha$.
\end{enumerate}

The \emph{bias detection criterion} used in production requires bootstrap confidence intervals for all demographic groups to overlap. Non-overlap constitutes evidence of statistically significant disparity. Severity is assessed by the magnitude of non-overlap and effect size.

\paragraph{Computational cost.}
The bootstrap multiplies HE aggregations by $B$; $B = 1000$ is typical for stable 95\% confidence intervals.
But each resample is independent, so all $B$ aggregations are fully parallelizable.
This makes the bootstrap practical even at production scale.

\section{BISG for Race \& Ethnicity Fairness Measurements: A Recap}
\label{sec:f-audits}
\label{sec:bisg}

We focus on race and ethnicity fairness measurements that can be recast as disaggregated evaluations~\cite{barocas2021} of AI systems: assessments and comparisons of system performance across race/ethnicity groups.
Generating such measurements relies on two data elements: member-level race/ethnicity data and member-level system performance data (typically: member identifier, ground truth, AI output).
Disparity estimators~\cite{chen2019, elzayn2023} aggregate these member-level elements into group-wise sample statistics.
In this section, we review the first element: the member-level race/ethnicity signal.

A key PPRE design goal was to avoid assigning individual members to categorical race/ethnicity groups.
Elliott et al.~\cite{elliott2009} further note that categorical assignment incurs more error than probabilistic estimation.
PPRE first relies on the BISG method to generate and supply probabilistic, non-categorical proxies for member-level race/ethnicity information.
This addresses the systemic problem of the absence of comprehensive race/ethnicity data.
BISG~\cite{elliott2009, voicu2018, imai2022} estimates an individual's probability $R$ of belonging to one of six exclusive race/ethnicity categories ($k=6$), using two inputs: surname $S$ and location $G$.
The estimate takes the form of a categorical posterior $R \, | \, S, G \sim Categorical(\vec{p} \; | \, S, G)$, computed via Bayes' theorem under a conditional independence assumption $S \perp G| R$:
\begin{equation}
    \mathsf{Pr}(r|s, g) = \frac{\mathsf{Pr}(r|s) \cdot \mathsf{Pr}(g|r)}{\sum_{r=1}^{k=6} \mathsf{Pr}(r|s) \cdot \mathsf{Pr}(g|r) } \;.
    \label{eq:bisg}
\end{equation}

Practitioners use the 2010 U.S.\ Census ``Frequently Occurring Surnames'' and ``Census Summary File'' tables~\cite{census_surnames_2016, census_summary_2012} to populate the $R|S$ and $G|R$ distributions, with location $G$ aggregated to ZIP Code Tabulation Areas (ZCTAs).
These tables follow U.S.\ federal standards~\cite{omb_1997} defining six race/ethnicity categories: White, Black, Hispanic or Latino, American Indian or Alaska Native, Asian or Pacific Islander, and Multiple.
Voicu~\cite{voicu2018} extends BISG to BIFSG by including first names as an additional input dimension; our implementation relies on only the public Census tables.
See~\cite{ppre-arxiv2024} for a detailed derivation.

This probabilistic, non-categorical output is what satisfies PPRE's no-individual-assignment principle while still enabling aggregate measurement.
The conditional independence assumption $S \perp G| R$ is a useful simplification but likely violated in practice: within a given race/ethnicity group, surname and location are often correlated~\cite{imai2022}.
Additional known limitations include age-related degradation~\cite{rieke2022imperfect}, downstream estimator bias~\cite{chen2019, kallus2022}, and narrow categorization~\cite{pai-demog2021}.
We adopt the six-category scheme because it is the most widespread standard in U.S.\ government and academic analyses and reporting.

Our reliance on BISG as a demographic signal source has two downstream implications for the PPRE system design.
First, BISG outputs are probabilistic proxies. 
Thus we need to generalize the standard disparity estimators to handle \emph{probabilistic} demographic attributes in place of the categorical attributes the estimators usually ingest.
\sectionref{audit-func} develops the probabilistic disparity estimator framework and applies it to specific fairness metrics we use.
Second, the BISG estimation process incurs some statistical error, which can propagate through to the final disparity estimates.
Refining and further calibrating BISG for the LinkedIn member population remains important future work to control this second concern.



\section{PPRE Fairness Metric Algorithms}
\label{sec:metrics}
\label{sec:audit-func}

The fairness metrics in our responsible AI portfolio include two broad classes of metrics: those that evaluate equal treatment in the system's behavior with respect to candidates (the individuals being scored or ranked), and those that evaluate equal treatment in the system's behavior with respect to viewers (the users consuming the scores or rankings).
We refer to these as \emph{candidate-side} and \emph{viewer-side} metrics, respectively.
Our candidate-side metrics include the Equal Revocation of Opportunity (ERO) metric (\sectionref{ero}) and the Listwise Outcome Test (LOT) (\sectionref{lot}); the latter extends the outcome-test methodology of Knowles, Persico, and Todd~\cite{knowles2001racial} and Simoiu et al.~\cite{simoiu2017problem} to ranked-list settings.
Viewer-side metrics include the minimum quality of service with respect to the NDCG performance metric (MQOS-NDCG) (\sectionref{mqos-ndcg}) amongst others.
In the standard reporting of these metrics to the AI governance process, we measure and report both the point estimates per demographic group and the statistical-significance for the point estimates.
Disparities are certified by the observation of statistically-significant metric differences across the probed demographic groups.

Standard disparity estimators rely on categorical indicators $\mathbb{I}(R_i = j)$ to assign each individual to exactly one demographic group.
PPRE uses probabilistic BISG estimates instead, so the estimators must be generalized.
Many fairness metrics we use in production are \emph{model performance disparity estimators}: they evaluate an AI system's behavior by comparing a function of ground-truth outcomes and model predictions across demographic groups.
The relevant categorical-attribute version of such a fairness estimator is the following standard sample mean estimator for group $j$:
\begin{equation}
    \hat{\mu}[j] = \frac{\sum_{i=1}^{n} \mathbb{I}(R_i=j) \cdot f(Y_i, \hat Y_i)}{\sum_{i=1}^{n} \mathbb{I}(R_i=j)}
\end{equation}
where $f(Y_i, \hat Y_i)$ is a model performance metric (a function of ground truth $Y_i$ and model prediction $\hat Y_i$), and $\mathbb{I}(R_i = j)$ is the indicator function of member $i$ belonging to group $j$.
Following Chen et al.~\cite{chen2019}, we replace hard indicators with soft indicators $\mathsf{Pr}(R_i = j)$.
Elzayn et al.~\cite{elzayn2023} establish the asymptotic properties of this replacement.
The general form is a weighted ratio:
\begin{equation}
    \mu[j] = \frac{\sum_{i=1}^{n} \mathsf{Pr}(R_i=j) \cdot f(Y_i, \hat Y_i)}{\sum_{i=1}^{n} \mathsf{Pr}(R_i=j)}, \; \forall j \in \{1, \ldots, k\}
    \label{eq:soft-ai-sample-mean}
\end{equation}
where $\mathsf{Pr}(R_i = j)$ is the BISG posterior probability that member $i$ belongs to group $j$.
Each member contributes fractionally to every group's statistic, in proportion to their posterior group-membership probability.

This estimator has two important properties.
First, it is linear in the test values $f(Y_i, \hat Y_i)$, so it is computable entirely within the additive HE scheme (\sectionref{pets}).
Second, the ratio form decomposes naturally into separate numerator and denominator aggregates, which is precisely the ratio estimation pattern from \sectionref{ratio-estimation}.
These two properties are what make encrypted fairness computation feasible.

The metrics fall into two families.
\emph{Candidate-side metrics} ask whether individuals from different demographic groups are treated equitably by the system; ERO (\sectionref{ero}) and LOT (\sectionref{lot}) belong to this family.
\emph{Viewer-side metrics} ask whether the quality of service delivered to users differs across demographic groups; MQOS-NDCG (\sectionref{mqos-ndcg}) belongs to this family.

The next subsections present the PPRE-adapted versions of the ERO, LOT, and MQOS-NDCG metrics, accounting for both probabilistic group-membership and AHE-computation.
The PPRE-adapted metrics rely on bootstrap hypothesis testing (\sectionref{bootstrap}) for significance testing.
Two additional metrics in the RAI portfolio, the Pointwise Outcome Test (POT) and MQOS-AUC, have been fully designed but are not yet deployed; their specifications appear in Appendix~\ref{sec:appendix-metrics}.
Appendix~\ref{sec:appendix-estimators} presents additional estimator types (output metric and probabilistic count) that extend the framework beyond model performance evaluation.

\subsection{ERO: Equal Revocation of Opportunity}
\label{sec:ero}

Suppose a scoring model flags candidates for further review. Some flagged candidates will be false positives: the model predicted a positive outcome that did not materialize. ERO asks whether the burden of these false flags falls equally across demographic groups, or whether some groups disproportionately experience incorrect positive predictions.

With categorical group labels, the group-$j$ false positive rate is 
\begin{equation}
    \text{FPR}[j] = \frac{\sum_i \mathbb{I}(R_i = j) \cdot \mathbb{I}(\hat{Y}_i = 1,\, Y_i = 0)}{\sum_i \mathbb{I}(R_i = j)} \;.
\end{equation}
ERO flags a disparity when $\max_j \text{FPR}[j] - \min_j \text{FPR}[j]$ exceeds a fairness threshold $\tau$.
The probabilistic adaptation is an instance of Equation~\ref{eq:soft-ai-sample-mean} with $f(Y_i, \hat{Y}_i) = \mathbb{I}(\hat{Y}_i = 1,\, Y_i = 0)$:
\begin{equation}
    \mu[j] = \frac{\sum_{i=1}^{n} \mathsf{Pr}(R_i = j) \cdot \mathbb{I}(\hat{Y}_i = 1,\, Y_i = 0)}{\sum_{i=1}^{n} \mathsf{Pr}(R_i = j)}, \quad \forall j \in \{1, \ldots, k\}.
    \label{eq:ero-prob}
\end{equation}
The HE computation is straightforward: $P_2$ computes the false-positive indicator in plaintext before encryption (pre-encryption comparison), and $P_1$ aggregates via the ratio estimation pattern.
ERO uses only these two adaptation patterns plus bootstrap hypothesis testing (Table~\ref{tab:metric-patterns}).

\subsection{LOT: Listwise Outcome Test}
\label{sec:lot}

Consider a ranked list of candidates produced by a recommendation system. LOT asks whether the quality transitions between adjacent ranks depend on the candidates' demographic group membership: \emph{If a member of group $a$ is ranked just above a member of group $b$, is the normalized relevance gap between them consistent with what we would observe if group membership were irrelevant to ranking?}

In the standard formulation, for each query, the ranking model produces an ordered list. 
The ground-truth relevance $Y_i$ of the candidate at each rank position $i$ is observed after the fact.
To make relevance scores comparable across queries of different lengths and quality, we normalize the relevance scores by the \emph{Ideal Discounted Cumulative Gain} for the query: $\tilde{Y}_i = Y_i / \text{IDCG}$.
This prenormalization captures the relationship between the ground-truth relevance ($Y$) and the model's ranking output ($\hat{Y}$), since the ranking determines which $Y_i$ appears at each position.

For each ordered pair of successively-ranked members of demographic groups $(a \gneq b)$ (i.e. member from group $a$ ranked higher than member of group $b$), LOT computes the expected normalized-relevance difference between adjacently ranked candidates.
This ordered adjacency scheme is equivalent to weighting the successive $\tilde{Y}$-difference by the product of the binary demographic indicator functions for groups $a,b$ (higher-ranked candidate $\in a$ and lower-ranked candidate $\in b$):
\begin{equation}
    \text{LOT}_{a,b} = \frac{\sum_{i=1}^{n-1} (\tilde{Y}_i - \tilde{Y}_{i+1}) \cdot \mathbb{I}({R_i = a} \cap {R_{i+1} = b})}{\sum_{i=1}^{n-1} \mathbb{I}({R_i = a} \cap {R_{i+1} = b})}
    \label{eq:lot}
\end{equation}
where $\tilde{Y}_i = Y_i / \text{IDCG}$ is the IDCG-normalized relevance at position $i$, $R_i$ is the race/ethnicity of the member at position $i$, and $\mathbb{I}(\cdot)$ is the indicator function.
The numerator accumulates normalized-relevance differences across all adjacent pairs of demographic groups, $a \gneq b$, while the denominator counts the total number of such adjacent pairs.
A $\text{LOT}_{a,b} \geq 0$ over all group pairs indicates that ranking quality transitions depend on relevance scores and are independent of demographic group.
Such a result means the ranking is fair with respect to demographics.

\paragraph{Probabilistic adaptation.}
The probabilistic LOT estimator replaces the hard indicators with BISG posteriors. 
Under an independence assumption between adjacent candidates, the joint group indicator becomes a product of posterior probabilities:
\begin{equation}
    \text{LOT}_{a,b} = \frac{\sum_{i=1}^{n-1} (\tilde{Y}_i - \tilde{Y}_{i+1}) \cdot \mathsf{Pr}(R_i = a) \cdot \mathsf{Pr}(R_{i+1} = b)}{\sum_{i=1}^{n-1} \mathsf{Pr}(R_i = a) \cdot \mathsf{Pr}(R_{i+1} = b)}
    \label{eq:lot-prob}
\end{equation}
The statistic retains the same ratio form, now with soft demographic weights.
Each adjacent pair contributes to every group-pair's LOT statistic in proportion to the product of the two members' posterior probabilities.

\paragraph{HE computation strategy.}
LOT's structure is well-suited to HE computation because the normalized-relevance differences and the demographic weights factor cleanly across the two parties.
$P_2$ holds outcomes and rankings, and can compute IDCG-normalized differences; $P_1$ holds race probabilities and can compute cross-group weights.
The computation uses three techniques from the extended HE toolkit: \textbf{pre-encryption comparison} (to precompute and presort the $\tilde{Y}$ differences in plaintext), \textbf{signed-number encoding} (to handle negative values in the $\tilde{Y}$ field), and \textbf{ratio estimation} (to compute the weighted average difference for each group pair).
Bootstrap significance testing (\sectionref{bootstrap}) operates on the decrypted LOT statistics, constructing confidence intervals and testing for statistically significant group-pair disparities.



\subsection{MQOS-NDCG: Minimum Quality of Service-NDCG}
\label{sec:mqos-ndcg}

LOT and ERO evaluate fairness from the candidate's perspective. But two-sided ranking markets also have viewers: the users who consume ranked lists. MQOS-NDCG evaluates fairness from the viewer's perspective, asking whether any demographic group experiences systematically lower ranking quality.
The metric checks whether any group's average NDCG falls below a threshold relative to the overall population.

For a query $q$, $\text{NDCG}(q) = \text{DCG}(q) / \text{IDCG}(q)$ where $\text{DCG}(q) = \sum_{p=1}^{n_q} (2^{\text{rel}(q,p)} - 1) / \log_2(p+1)$ and IDCG is the DCG of the ideal ranking.
MQOS-NDCG flags a disparity when $\overline{\text{NDCG}}_{\text{overall}} - \overline{\text{NDCG}}[j] > \tau$ for any group $j$.
The probabilistic adaptation weights each query's NDCG by the issuing member's posterior:
\begin{equation}
    \overline{\text{NDCG}}[j] = \frac{\sum_{q} \mathsf{Pr}(R_q = j) \cdot \text{NDCG}(q)}{\sum_{q} \mathsf{Pr}(R_q = j)}
    \label{eq:ndcg-prob}
\end{equation}
This is an instance of Equation~\ref{eq:soft-ai-sample-mean} with $f$ being the pre-computed NDCG value for each query.
The HE computation follows the same pattern as ERO: $P_2$ computes per-query NDCG in plaintext before encryption (eliminating division from the HE path), and $P_1$ aggregates via ratio estimation with random masking.
Unlike LOT, MQOS-NDCG involves no signed differences and no cross-member interactions (Table~\ref{tab:metric-patterns}).

\subsection{Summary of Adaptation Patterns by Metric}
\label{sec:metric-summary}

Table~\ref{tab:metric-patterns} summarizes which HE toolkit extensions each metric requires.
The pattern is instructive: ratio estimation and fixed-point encoding are universal because every metric is a weighted ratio involving real-valued BISG probabilities encoded as Paillier integers.
Pre-encryption comparison is also universal, since every metric involves at least one operation (thresholding, sorting, or differencing) that $P_2$ performs in plaintext before encryption.
Signed encoding is needed only for LOT, where score differences can be negative.

\begin{table*}[h]
\centering
\caption{HE adaptation patterns used by each PPRE fairness metric.}
\label{tab:metric-patterns}
\begin{tabular}{lccccc}
\toprule
\textbf{Metric} & \textbf{Ratio Est.} & \textbf{Fixed-Pt.} & \textbf{Pre-Enc. Cmp.} & \textbf{Signed Enc.} & \textbf{Bootstrap} \\
\midrule
ERO         & \checkmark & \checkmark & \checkmark &            & \checkmark \\
LOT         & \checkmark & \checkmark & \checkmark & \checkmark & \checkmark \\
MQOS-NDCG   & \checkmark & \checkmark & \checkmark &            & \checkmark \\
\midrule
POT$^\dagger$        & \checkmark & \checkmark & \checkmark &            & \checkmark \\
MQOS-AUC$^\dagger$   & \checkmark & \checkmark & \checkmark &            & \checkmark \\
\bottomrule
\multicolumn{6}{l}{\footnotesize $^\dagger$Designed but not yet deployed in production. See Appendix~\ref{sec:appendix-metrics}.}
\end{tabular}
\end{table*}

\section{The PPRE System: Composing the Protocol}
\label{sec:ppre-system}
\begin{figure*}[ht!]
\centering
\includegraphics[width=\textwidth]{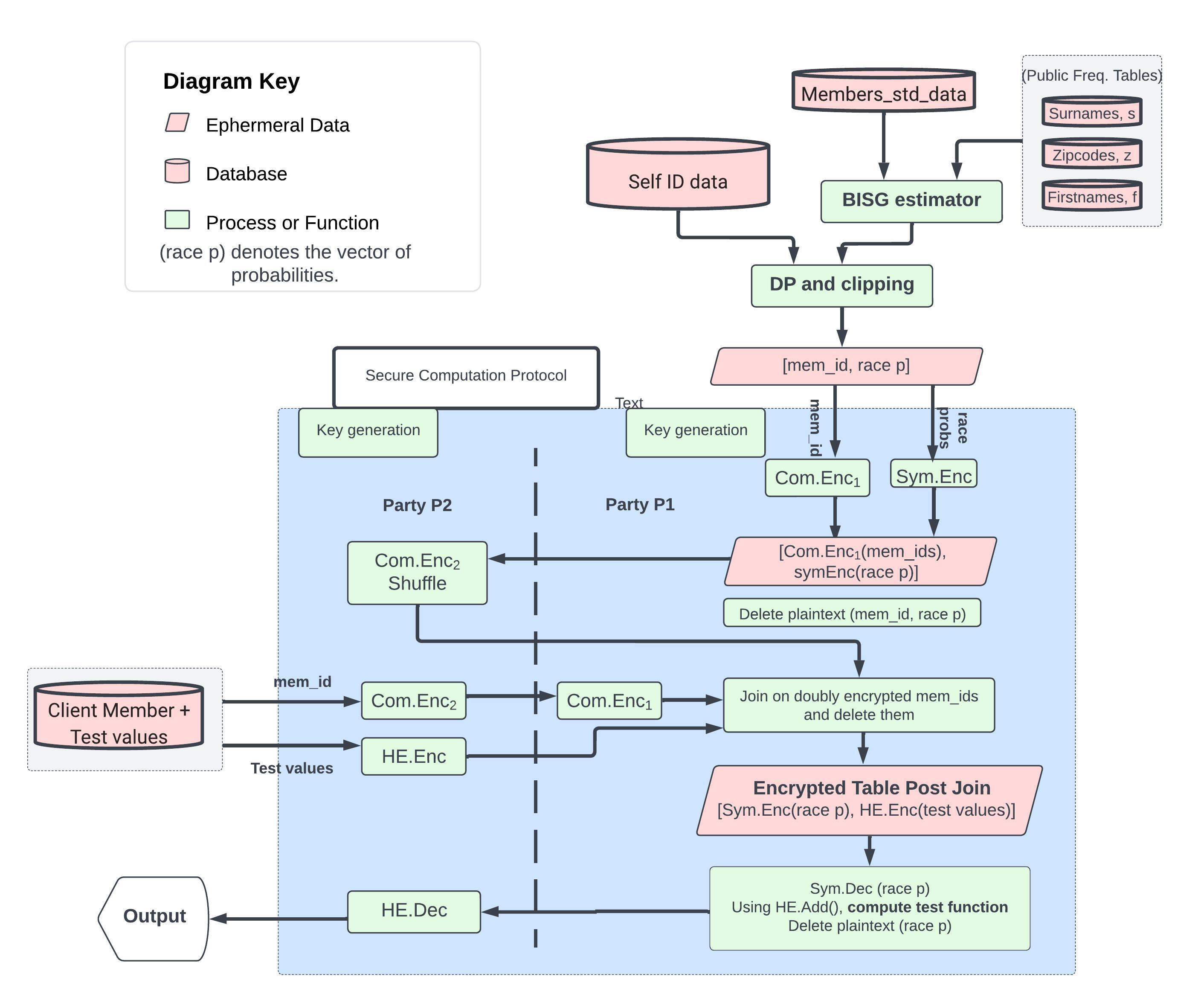}
\caption{PPRE system}
\label{fig:ppre-system}
\end{figure*}

This section assembles an end-to-end two-party computation protocol from the components developed in the preceding sections: the demographic signal processing (\sectionref{f-audits}, \sectionref{metrics}) and the privacy-primitives toolkit (\sectionref{pets}, \sectionref{he-adapt}).
Two entities participate: (a) the tester ($P_1$), the data privacy/responsible AI team, and (b) the test client ($P_2$), the product owner whose AI model is under test. \figureref{ppre-system} depicts the overall system.

At a high level, the protocol has two flavors of sub-processes.
$P_1$ first computes a privacy-preserving probabilistic race/ethnicity dataset using BISG and Self-ID. 
This is primarily a signal-processing sub-process.
The two parties then execute a secure computation protocol, built from the cryptographic primitives and their HE extensions, to join their private datasets and evaluate test functions that produce aggregated outputs.
This is a PET-based secure computation sub-process.
The test functions (\sectionref{audit-func}) are weighted sample means: linear operations over $P_2$'s test values weighted by $P_1$'s race/ethnicity probabilities. Additive homomorphic encryption fully supports this class of computation.
This is primarily a signal-processing sub-process, albeit in ciphertext-space.

\subsection{Data Preparation and Local DP}
\label{sec:clipping}
Before the secure computation begins, $P_1$ must construct its demographic dataset.
This involves combining two data sources (BISG and Self-ID) and applying privacy protections to prevent re-identification.

First, $P_1$ applies local DP (randomized response) to Self-ID data. Each Self-ID record is a one-hot vector over $k=6$ race/ethnicity categories, e.g.\ $(1,0,0,0,0,0)$. Randomized response retains the true value with probability $\frac{e^\epsilon}{(k-1+e^\epsilon)}$ and flips it to one of the other $(k-1)$ entries with probability $\frac{1}{(k-1+e^\epsilon)}$ each; $\epsilon$ is tunable.

Next, $P_1$ computes BISG estimates and combines them with the local-DP Self-ID table. The combined dataset then undergoes \emph{probabilistic clipping}: when a race/ethnicity probability exceeds a threshold $\bT$, $P_1$ reduces it by a small uniformly sampled random value and adjusts the remaining categories using a properly-scaled Dirichlet draw. $P_1$ sets $\bT$ so that $90\%$ of the largest BISG probability values fall below it.
This clipping serves a specific purpose: it prevents an adversary from distinguishing BISG-estimated records from Self-ID records by inspecting probability values.
In the flow diagram, $\text{(race p)}$ denotes the probability vector $\racep$.

\subsection{Secure Computation}
\label{sec:secure-computation}
After data preparation, each party holds a private table.
$P_1$ holds rows $(\memid, \racep)$: member ID and six race/ethnicity probabilities.
$P_2$ holds rows $(\memid, \values)$: member ID and test values.
The protocol's task is to join these tables on member ID, compute weighted aggregates, and reveal only the final aggregate statistics.
Neither party should learn the other's plaintext values at any point.

We denote by $\G$ the group for commutative encryption and by $\RO: \{0,1\}^* \rightarrow \G$ a random oracle.
All probability values are scaled to integers for the HE and symmetric encryption schemes.

The protocol, inspired by PSI-Sum~\cite{IKNPRSSSY19} and circuit PSI~\cite{PSWW18}, proceeds through seven steps:
\begin{itemize}
	\item \textbf{Key generation.} In practice, this occurs in parallel with data preparation.
		\begin{itemize}
		\item Instantiate a fresh random oracle $\RO$.
		\item $P_1$ generates $\sk_1 \leftarrow \comkeygen(1^\secr)$ and $\symsk \leftarrow \symkeygen(1^\secr)$.
		\item $P_2$ generates $\sk_2 \leftarrow \comkeygen(1^\secr)$ and $(\ahepk,\ahesk) \leftarrow \ahekeygen(1^\secr)$.
		\end{itemize}
	Each party's keys are generated locally and held in memory only (\sectionref{key-management}).

	\item \textbf{Encrypt $P_1$ data.} $P_1$ encrypts member IDs with commutative encryption and race probabilities with symmetric encryption, sending $\{(\comenc_{\sk_1}(\RO(\memid)), \symenc_\symsk(\racep; \rand_i))\}$ to $P_2$. $P_1$ then deletes its plaintext dataset $\{\memid, \racep\}$. From this point forward, $P_1$ holds no plaintext demographic data.

	\item \textbf{Doubly encrypt $P_1$ data.} $P_2$ applies its commutative key to compute
	\[ \{(\comenc_{\sk_2}(\comenc_{\sk_1}(\RO(\memid))), \symenc_\symsk(\racep; \rand_i))\}, \]
	then shuffles the dataset and sends it back to $P_1$. The shuffle is critical: it breaks the positional correspondence between $P_1$'s original records and the returned records, preventing $P_1$ from inferring which members are in $P_2$'s dataset.

	\item \textbf{Encrypt $P_2$ data.} $P_2$ encrypts its member IDs with commutative encryption and test values with additive HE, sending $\{(\comenc_{\sk_2}(\RO(\memid)), \aheenc_\ahepk(\values; \rand_j))\}$ to $P_1$.

	\item \textbf{Privacy-preserving join.} $P_1$ doubly encrypts $P_2$'s member IDs using $\sk_1$ and joins the two datasets on the doubly encrypted identifiers. Commutativity guarantees
	\[ \comenc_{\sk_1}(\comenc_{\sk_2}(\RO(\memid))) = \comenc_{\sk_2}(\comenc_{\sk_1}(\RO(\memid))). \]
	$P_1$ then deletes all doubly encrypted identifiers, producing the de-identified joined table
	\[ \{(\symenc_\symsk(\racep; \rand_i), \aheenc_\ahepk(\values; \rand_j))\}. \]
	At this point, neither party can link any row in the joined table back to a specific member.

	\item \textbf{Test computation.} $P_1$ decrypts the race probabilities using $\symsk$ to obtain
	\[ \{(\racep, \aheenc_\ahepk(\values; \rand_j))\}. \]
	The test functions (\sectionref{audit-func}) are linear combinations of $\{\values\}$ weighted by $\{\racep\}$. $P_1$ therefore applies $\aheadd$ to compute the encrypted aggregates
	\[ \{(\mathsf{race_i}, \aheenc_\ahepk(\out_i, r_i))\}_{i=1}^{6} \]
	and sends them to $P_2$.
	This is the core computation step: $P_1$ can weight $P_2$'s encrypted test values by plaintext race probabilities without ever decrypting those test values.

	\item \textbf{$P_2$ learns output.} $P_2$ decrypts with $\ahesk$ to learn the aggregate output. When only $P_1$ should learn the output, $P_1$ masks the encrypted aggregates before transfer. $P_2$ decrypts and returns the masked values, learning nothing, and $P_1$ removes the mask.
\end{itemize}

\subsection{Privacy and Security Guarantees}
\label{sec:privacy-guarantees}
The composition of these protocol steps produces a set of emergent privacy guarantees that no single step provides alone:
\begin{enumerate}
    \item \textbf{No party learns complete sensitive data.} No single party ever simultaneously holds the plaintext tuple $(\memid, \racep, \values)$ for any member. $P_1$ sees race probabilities but never test values in plaintext; $P_2$ sees test values and aggregate statistics but never individual race probabilities.
    \item \textbf{No identifiable plaintext analysis.} All measurement computations run on cryptographically de-identified records. $P_1$ deletes the plaintext table $\{(\memid, \racep)\}$ immediately after encryption.
    \item \textbf{Exchange of encrypted-only data.} Every value exchanged between parties is encrypted under keys the receiving party does not hold. $P_2$ cannot decrypt the symmetrically encrypted race probabilities; $P_1$ cannot decrypt the homomorphically encrypted test values.
    \item \textbf{Shuffling prevents re-identification by ordering.} $P_2$'s random permutation breaks the positional correspondence between $P_1$'s original records and the doubly-encrypted records returned. This prevents $P_1$ from determining which of its rows participated in the join.
    \item \textbf{Multi-layer re-identification defense.} Local differential privacy on Self-ID records, probabilistic clipping on probability vectors, and governance controls together reduce re-identification risk across multiple measurement sessions.
    \item \textbf{Doubly-encrypted IDs are discarded.} All doubly-encrypted member identifiers are deleted before the measurement calculation begins. No linkable identifiers persist through the computation phase.
\end{enumerate}
These guarantees are architectural, not parametric: they hold regardless of the specific fairness metric being computed or the size of the dataset.

\section{Production Deployment and Evaluation}
\label{sec:impl}
This section describes the production instantiation of PPRE: the cryptographic and privacy parameter choices that govern the system's security-utility tradeoff, and the deployment architecture that hosts it.

\subsection{Cryptographic and Privacy Parameters}
\label{sec:params}
The cryptographic parameters reflect standard choices for 128-bit security.
Table~\ref{tab:crypto-params} summarizes them.
\begin{table*}[ht!]
\centering
\begin{tabular}{l l l}
\toprule
\textbf{Parameter} & \textbf{Choice} & \textbf{Details} \\
\midrule
Security parameter $\secr$ & $128$ & Standard 128-bit security level \\
Commutative encryption & Pohlig-Hellman on Curve25519 & \cite{Ber06} \\
Random oracle & SHA-256 & Hash-into-group via rejection sampling \\
Additive HE & Paillier cryptosystem & 2048-bit modulus~\cite{Paillier99} \\
Symmetric encryption & AES-256 & 256-bit key \\
\bottomrule
\end{tabular}
\caption{Cryptographic parameters for the PPRE system.}
\label{tab:crypto-params}
\end{table*}

The privacy parameters control the tradeoff between demographic data fidelity and re-identification risk.
Table~\ref{tab:dp-params} summarizes the choices applied during data preparation (\sectionref{clipping}).
A local DP $\epsilon$ of $4.5$ gives a flipping rate of $5.26\%$.
Taken in isolation, $\epsilon = 4.5$ is on the high end of what the DP literature considers a strong standalone guarantee ($e^{4.5} \approx 90$).
In PPRE, however, local DP is one layer in a defense-in-depth architecture.
The combination of $\epsilon$-local DP on Self-ID records, probabilistic clipping that prevents near-certain assignments, ephemeral computation with immediate deletion of member-level data, key destruction after each session, and governance controls that enforce minimum population thresholds collectively provides a protection profile substantially stronger than any single mechanism alone.
Local DP's primary role in this composition is to prevent re-identification of Self-ID respondents across repeated measurement runs; the cryptographic and architectural guarantees (\sectionref{privacy-guarantees}) carry the bulk of the within-session privacy protection.
The clipping threshold $\bT = 0.825$ is set so that $90\%$ of the largest BISG probability values fall below it, ensuring that probabilistic clipping affects only the most confident estimates.
\begin{table}[ht!]
\centering
\begin{tabular}{l l l}
\toprule
\textbf{Parameter} & \textbf{Value} & \textbf{Notes} \\
\midrule
Local DP $\epsilon$ & $4.5$ & Flipping rate $= 5.26\%$ \\
Clipping threshold $\bT$ & $0.825$ & $90\%$ of largest BISG probabilities fall below $\bT$ \\
\bottomrule
\end{tabular}
\caption{Differential privacy parameters for the PPRE system.}
\label{tab:dp-params}
\end{table}

\subsection{System Architecture}
\label{sec:architecture}
The two parties execute their protocol steps as independent jobs on a distributed compute cluster.
Because PPRE operates in batch mode, the parties exchange data asynchronously through a shared distributed file system rather than a direct network channel.
Only encrypted data is stored at shared locations; plaintext data and keys remain exclusively in each party's job memory.

A read-write lock scheme coordinates protocol steps, and access control on the shared file system enforces strict isolation: each party can only access its own designated directory.
This architecture requires no dedicated communication infrastructure beyond what the existing compute platform provides.


\subsection{Production Fairness Measurements}
\label{sec:audit-results}

We present production fairness measurements of two AI ranking systems.
The (LOT, MQOS-NDCG) measurement pair covers the two key perspectives in two-sided ranking applications: LOT evaluates candidate-side fairness (whether HSM members are under- or over-ranked), and MQOS-NDCG evaluates viewer-side fairness (whether each group experiences similar quality of service).
All tests use bootstrap hypothesis testing in ciphertext space.
The severity criterion requires confidence intervals for all demographic groups to overlap; non-overlap constitutes evidence of statistically significant disparity.

\subsubsection{Candidate-side Fairness Measurement for a Member Ranking System}
\label{sec:audit-rs}

We evaluated an AI system that ranks members for a one-sided ranking market using the LOT (\sectionref{lot}).
For this evaluation, we coarsened the race/ethnicity categories to a two-class probabilistic split: HSM vs. non-HSM split, where HSM (historically marginalized) includes Black, Hispanic, and Native American groups.
The system showed {low bias}: none of the top six ranks showed HSM members under- or over-ranked relative to non-HSM members in nearby ranks.
Confidence intervals for outcome differences between subsequent (HSM, non-HSM) and (non-HSM, HSM) rank pairs did not reach statistical significance in the negative direction.
A negative difference would indicate the lower-ranked item has measurably higher ground-truth relevance, signaling unjustified ranking based on HSM status.
Because this system is a one-sided, candidates-only market, the MQOS-NDCG viewer-side test is not meaningful here.

\subsubsection{Viewer-side Fairness Measurement for an Item Recommendation System}
\label{sec:audit-jymbii}

The next system-under-measurement is a ranking model that proactively recommends items to viewing members through multiple viewing channels.
We use a viewer-side measurement, MQOS-NDCG, since this system serves a one-sided viewer market.
For this measurement, we do not collapse the $6$ race/ethnicity classes defined by the U.S.\ Office of Management and Budget (OMB)~\cite{omb_1997} into HSM-vs-nonHSM. 

The PPRE-MQOS-NDCG measurement on the system shows overlapping confidence intervals for all groups, indicating no statistically significant differences at $96\%$ confidence.
This measurement officially falls into the low bias severity criterion.
However, all group metrics fall below $80\%$ of the overall test dataset's NDCG.
This suggests the PPRE test subsample differs from the system's full test population.
The most likely cause is the geofencing needed to restrict the PPRE measurement to a subset of the USA member population.


\section{Discussion}
\label{sec:disc}

The PPRE system, as built, represents an existence proof that it is feasible to do fairness measurements under strong privacy guarantees at industrial scale.
PPRE now powers race/ethnicity fairness measurements for the platform.
Besides the final outcome, there are useful lessons to draw from our process of conceptualizing, designing, and implementing the system.
In this section, we will examine and discuss sample runs of the PPRE system on large-scale AI systems, distill operational learnings, identify limitations, and highlight open questions.

\subsection{Examining Sample PPRE Measurements}
\label{sec:disc-results}

The production fairness measurements (\sectionref{audit-results}) found low bias across the systems tested.
But the numbers themselves tell only part of the story.
All intermediate computations occur in ciphertext space; the only plaintext values $P_2$ ever observes are the final per-group aggregate statistics and their bootstrap confidence intervals.
Several aspects of how these results are tested for significance and interpreted merit careful discussion.

\subsubsection{Measurement Quality Discussion}
\label{sec:measurement-quality}
The (LOT, MQOS-NDCG) pair covers the two key perspectives in two-sided ranking.
LOT evaluates whether candidates are systematically under- or over-ranked by group; MQOS-NDCG evaluates whether viewers from each group experience comparable ranking quality.
Together, they can provide a more complete assessment of fairness for two-sided recommendation markets.

The candidate-side measurement above featured a condensation of the race/ethnicity signal into a dichotomous probability signal, HSM vs non-HSM.
The primary motivation for this condensation is the data needs of the LOT test. 
LOT requires finding ranking pairs that feature the different demographic attributes in the ranked slots.
A $6$-dimensional attribute signal would require $\binom{6}{2}$ such pairs to be represented well-enough in the test dataset.
The $2$-dimensional condensation allows us to drop that required count down to just $2$ pairs.
This yields a statistical test with higher discriminatory power for the condensed version of the test hypothesis. 
Furthermore, the condensation is semantically-relevant for a fairness measurement; it still compares system behavior between a group with historic patterns of disadvantage vs. a group without those historic patterns. 
Future work can aim to expand the group-specificity of the test while keeping the test's power high.

\subsubsection{Geofencing effects on measurement populations.}
The viewer-side measurement above highlights a general challenge for measurement systems that depend on geographically-scoped demographic models.
Records from non-USA members need to be excluded because BISG is calibrated for the U.S.\ population.
Furthermore, records for residents of certain USA states are also excluded due to local privacy laws.
These exclusions alter the test population relative to the system's full actual user base.
These exclusions must be factored into contextualization discussions of PPRE measurement results.
For example, these exclusions inform our choice to base severity criteria mainly on CI overlap; the comparison to overall NDCG is provided as informative context.
These baked-in selection effects do not invalidate the fairness comparison within the included population, but they limit generalizability.
Quantifying the impact of such exclusions on measurement representativeness is an important direction for future work.

\subsubsection{Significance testing.}
A key question for any fairness measurement is whether observed group differences are statistically meaningful or merely sampling noise.
PPRE addresses this entirely through bootstrap hypothesis testing (\sectionref{bootstrap}).
Each bootstrap resample produces a fresh set of encrypted numerator-denominator pairs, which P1 aggregates and P2 decrypts independently.
From $B$ resampled statistics, P2 constructs percentile confidence intervals for each group's metric.
The bias detection criterion requires overlap of confidence intervals across all demographic groups.
Non-overlap at a specified confidence level (96\% in the job recommendation measurement; 95\% elsewhere) constitutes evidence of statistically significant disparity.
This approach avoids parametric distributional assumptions that would be difficult to verify for the weighted-ratio statistics PPRE computes.
And it avoids the need for squaring or variance computation in ciphertext space, neither of which additive HE supports.
A practical consequence is that the bootstrap multiplies the HE aggregation workload by $B$.
But each resample is independent, so all $B$ aggregations are fully parallelizable.
In production, $B = 1000$ yields stable 95\% confidence intervals.


\subsubsection{Caveats on BISG's Calibration Quality.}
The BISG validation (Appendix~\ref{sec:bisg-validation}) compared three estimation variants against Self-ID survey responses using cross-entropy.
Baseline BISG (surname plus geolocation) performed best overall.
Adding first-name distributions (BIFSG) yielded no meaningful improvement, while dropping geolocation (BISF) slightly improved aggregate cross-entropy at the cost of degraded performance for smaller groups.
The implication is clear: geolocation provides critical signal for geographically concentrated groups, and removing it trades aggregate accuracy for worse tail-group estimates.

These results guided a concrete design choice: baseline BISG is used for all production measurements.
But two caveats temper this conclusion.
First, the Self-ID survey is self-selected, covering roughly $6\%$ of U.S.\ members.
The comparison measures BISG accuracy only on the subset of members who chose to respond, and that subset may not be representative of the broader population.
Second, cross-entropy measures calibration of the full probability vector, not accuracy of a point assignment.
A well-calibrated BISG model can still produce posterior distributions that are diffuse for members whose surnames and locations are not strongly associated with any single group.
Such diffuse posteriors are handled correctly by the soft-indicator estimators (\sectionref{audit-func}), which weight each member's contribution proportionally rather than forcing a categorical assignment.

\subsection{Validating HE Metric Transformations}
\label{sec:disc-validation}

Computing fairness metrics under homomorphic encryption introduces a translation layer: each metric's standard definition must be decomposed into operations that additive HE supports.
The question is whether this decomposition preserves the metric's semantics.
Errors in the translation would silently corrupt results, and the encrypted computation provides no diagnostic visibility.
We validated the most complex translation, the LOT metric, using controlled synthetic data.

\subsubsection{LOT Validation via Synthetic Data}
\label{sec:lot-validation}

The LOT statistic (Equation~\ref{eq:lot}) involves signed score differences, cross-group probability weights, and ratio estimation.
It exercises four of the five HE adaptation patterns simultaneously: ratio estimation, fixed-point encoding, pre-encryption comparison, and signed-number encoding.
This makes LOT the most demanding metric to validate, and the most informative: if the HE translation preserves correctness for LOT, the simpler metrics (ERO, MQOS-NDCG) are covered as well, since they use strict subsets of these adaptations (Table~\ref{tab:metric-patterns}).

We constructed a synthetic dataset where the ground-truth LOT values are known by design.
The key idea is to make race/ethnicity probabilities statistically independent of outcomes, so that the LOT statistic computed on the data should match predefined score gaps for every group pair:
\begin{enumerate}
    \item Generate a BISG table with random race/ethnicity probability vectors for each synthetic member.
    \item For each query, define ground-truth score gaps between adjacent ranks. These gaps are the target LOT values.
    \item Generate outcomes for the first-ranked item, then derive outcomes at subsequent ranks using the predefined gaps plus small additive noise.
    \item Generate model scores as a monotonically decreasing sequence (e.g., $0.9, 0.8, \ldots, 0.0$ for 10 ranks).
    \item Repeat for 40{,}000 queries to ensure statistical stability.
\end{enumerate}

We ran the full PPRE-LOT pipeline on the synthetic dataset: encryption, private set intersection, HE aggregation with bootstrap resampling, decryption, and final statistic computation.
Table~\ref{tab:lot-validation} shows representative results.

\begin{table}[ht!]
\centering
\caption{LOT validation: HE-computed values vs.\ synthetic ground truth. Each row is one adjacent-rank pair across two demographic groups. Standard deviations are from $B = 1{,}000$ bootstrap resamples.}
\label{tab:lot-validation}
\begin{tabular}{lcc}
\toprule
\textbf{Rank pair} & \textbf{Ground truth} & \textbf{PPRE-LOT (HE)} \\
\midrule
$r_{1\text{-}2}$ & $0.1200$ & $0.1202 \pm 1.5 \times 10^{-4}$ \\
$r_{2\text{-}3}$ & $0.3400$ & $0.3397 \pm 1.5 \times 10^{-4}$ \\
$r_{3\text{-}4}$ & $-0.2700$ & $-0.2698 \pm 1.4 \times 10^{-4}$ \\
$r_{4\text{-}5}$ & $0.7800$ & $0.7800 \pm 1.4 \times 10^{-4}$ \\
$r_{5\text{-}6}$ & $-0.4300$ & $-0.4301 \pm 1.5 \times 10^{-4}$ \\
$r_{6\text{-}7}$ & $-0.2400$ & $-0.2401 \pm 1.5 \times 10^{-4}$ \\
$r_{7\text{-}8}$ & $-0.2900$ & $-0.2900 \pm 1.4 \times 10^{-4}$ \\
$r_{8\text{-}9}$ & $0.7600$ & $0.7600 \pm 1.5 \times 10^{-4}$ \\
$r_{9\text{-}10}$ & $-0.4100$ & $-0.4100 \pm 1.6 \times 10^{-4}$ \\
\bottomrule
\end{tabular}
\end{table}

The HE-computed LOT values match the ground truth to within $3 \times 10^{-4}$ in all cases.
The residual discrepancy is attributable to the small additive noise injected during data generation and to fixed-point encoding precision.
Bootstrap standard deviations on the order of $10^{-4}$ confirm that the resampling procedure produces tight and stable confidence intervals at $B = 1000$.
And signed score differences (the negative entries in the table) are handled correctly by the signed-number encoding (\sectionref{signed-encoding}), confirming that the plaintext-space partitioning scheme works as designed.

This validation confirms that the full chain of HE adaptations preserves metric correctness for LOT.
Because ERO and MQOS-NDCG use strict subsets of these adaptations, the LOT validation provides coverage for those metrics as well.

\subsection{Operational Learnings}
\label{sec:disc-operational}

PPRE serves as an existence proof that privatized fairness measurement is feasible at industrial scale, with expected but manageable complexity costs.
Several operational learnings from the production deployment are worth distilling for institutions considering similar systems.

\paragraph{Testing dominates development cost.}
The hardest part of building PPRE was not the cryptographic protocol design.
It was testing and debugging multiparty computational flows.
Standard unit tests verify individual functions in isolation, but multiparty protocols introduce a class of bugs that arise from interactions between parties: timing, coordination, data exchange formatting, and encryption round-trip correctness.
These bugs are invisible to single-party tests.

Testing costs for multiparty flows exceeded those for comparable single-party implementations by at least an order of magnitude.
Essential supplements to unit testing included: end-to-end integration tests exercising the full protocol across both parties, protocol-level correctness checks verifying that encryption-decryption round-trips produce correct aggregates, coordination state-machine tests checking for race conditions in the asynchronous data exchange, and random-masking cancellation tests confirming that numerator-denominator mask factors cancel exactly after decryption.
Any institution building a similar system should budget accordingly.

\paragraph{Dual challenge: theory and engineering.}
PPRE's difficulty divides into two distinct components that reinforce each other.
The theoretical challenge is reformulating fairness metrics for computation under homomorphic encryption with probabilistic group memberships. This requires decomposing each metric into operations that additive HE supports while preserving statistical validity.
The engineering challenge is coordinating continuously communicating multiparty flows within pre-existing distributed infrastructure.
Both are substantial. Treating either as trivial given the other leads to underestimation of total effort.


\subsection{Transferable Concepts from this Work}
\label{sec:disc-framework}

The PPRE design pattern generalizes well beyond its current deployment context.
Any institution with (a) a sensitive demographic attribute it cannot freely store or use, (b) a disaggregated evaluation measurement it wants to compute, and (c) two or more internal entities holding different data components can apply the same architectural pattern.
The three-pillar structure maps naturally to transferable design principles, and the pillars are independently adoptable.

The PETs elements can be streamlined to just the PSI methods in the context of lightweight third-party measurements.
However, the AHE portion of the PETs stack does a lot of heavy lifting for multi-party measurements in a zero-trust setting where we do not want to maintain the assumption of a trustworthy party-1 or party-2 in the protocol.
The expectation should be that any application of the elements laid out in this work will need to have a detailed contextual understanding of the trust configuration among the parties, the regulatory mandates in play, the social context relevant to the fairness measurement, and the computational capabilities available.

Aspects of the demographic signal-processing can be swapped with other algorithms or sources that may be more applicable to a bias evaluation, e.g. in the E.U. geographic context.
In our use, probabilistic proxies replace categorical assignment, and soft-indicator estimators generalize standard fairness metrics to operate on probability distributions rather than categorical labels.
Our choices highlight a few salient degrees of freedom for other institutions exploring such a measurement process.
For one, the choice to work with probabilistic demographic proxies throughout.
Depending on the sensitivity of attribute in-focus, converting probabilistic proxies to categorical signal or working with categorical demographic signals \emph{ab initio} may be a useful modification; it would make the measurement algorithms somewhat simpler.
It is also possible to adopt key portions of PPRE's demographic signal processing independently of the cryptographic layer we used.
Furthermore, for institutions without a sparse golden demographic survey, the application of local DP application to the demographic signal source is moot.

There is a specific jurisdictional detail that enables our approach. 
The USA is a large-population jurisdiction with a uniform, standardized measurement concept for race/ethnicity. 
The OMB sets a standardized definition of the sociologically squishy concept of ``race/ethnicity'' with the $6$-dim categorization, codified in its 1997 memorandum~\cite{omb_1997}.
This standardized concept is reinforced by the US Census reliance on the OMB definitions for scoping the race/ethnicity survey questions as part of its constitutionally-mandated decennial Census.
The widespread reliance on the OMB-defined ``race/ethnicity'' construct does not mean the construct is perfect or above reproach. 
There are salient contestations regarding, for example, how people of Middle Eastern descent are meant to self-identify in the incomplete $6$-dimensional concept space.\footnote{This standard has since been revised. The OMB's 2024 update to Statistical Policy Directive No.~15~\cite{omb_2024} adds a \emph{Middle Eastern or North African} minimum category---directly addressing the self-identification gap noted here---and combines the race and ethnicity questions into a single co-equal item. Federal agencies must adopt the revised standard by March 2029, making it the appropriate target for future measurement deployments.}
Conceptual shortcomings notwithstanding, its standardization and widespread use mean our use of the same concept is legible to a broad cross-section of researchers, decision-makers, and laypersons. 
Other jurisdictions may not have the luxury of such a broadly-shared conception of ethnicity. 
The EU, for example, has different salient ethnicity concepts that are often only regionally salient to historical systematic disadvantage.
Consider the differential relevance of migration status, religion, Roma descent, and African descent across the EU member states; Farkas's comparative review~\cite{farkas2017analysis} documents both the variation in collection practices and the definitional inconsistency across jurisdictions.
These kinds of conceptual uncertainty can easily render a measurement effort foundationally unsound, no matter how much ingenuity goes into the technical measurement stack.
After all, what is the value of a fairness measurement relative to a demographic attribute upon which both decision-makers and stakeholders have no mutually settled definition?

From the privacy engineering contributions, our HE adaptations catalog (\sectionref{he-adapt}) is also highly transferable for anyone aiming to develop a privatized analytic capability that relies on additive homomorphic encryption.
Ratio estimation, fixed-point encoding, pre-encryption comparison, signed-number encoding, and bootstrap hypothesis testing apply to any ratio-form statistic computed under additive HE.
These patterns are not specific to fairness metrics or to race/ethnicity measurement; they address general limitations of additive homomorphic encryption that arise in any applied setting.

From the composed protocol, the two-party architecture with PSI-based join and homomorphic aggregation provides a reusable template for secure measurement protocols.
Governance controls (legal approval, minimum population thresholds, differencing-attack prevention) are separable from the cryptographic protocol and can be adapted to other institutional contexts.


\subsection{Limitations}
\label{sec:disc-limitations}

\begin{enumerate}
  \item \textbf{Geographical scope.}
    BISG depends on U.S.\ Census surname and geolocation frequency tables.
    Extension to the EU or other jurisdictions requires similarly well-curated census data with well-defined demographic groupings.
    Recent explorations of EU equality data suggest these standards are difficult to replicate outside the U.S., which constrains the framework's near-term international applicability.

  \item \textbf{BISG calibration.}
    The BISG model is not fine-tuned for the target member population.
    Validation confirms it is informative-but-imperfect for our target population.
    Calibration shortfalls in the demographic estimation propagate to downstream measurements.
    The Self-ID comparison characterizes accuracy only for Self-ID respondents, who measurably and systematically differ from non-respondents.

  \item \textbf{Diffuse posteriors.}
    For members whose surnames and locations are not strongly associated with any single group, BISG produces diffuse posteriors (e.g., near-uniform across categories).
    These members contribute nearly equally to every group's statistic, diluting the measurement signal.
    If a test population has many such members, measured disparities will be attenuated toward zero.

  \item \textbf{Geofencing and population selection.}
    Excluding non-USA members and certain state residents alters the test population.
    Fairness comparisons within the included population remain valid, but they may not generalize to the excluded population.
    Quantifying the impact of such exclusions on measurement representativeness is an important open problem.

  \item \textbf{LOT independence assumption.}
    The probabilistic LOT estimator (Equation~\ref{eq:lot-prob}) replaces the joint group indicator $\mathbb{I}({R_i = a} \cap {R_{i+1} = b})$ with the product $\mathsf{Pr}(R_i = a) \cdot \mathsf{Pr}(R_{i+1} = b)$, assuming independence between adjacent candidates' demographics.
    In practice, adjacent candidates in a ranked list may share characteristics because the ranking model clusters similar candidates.
    BISG posteriors depend on geolocation so spatial clustering can induce positive correlation between adjacent posteriors and possibly deflate the denominator for cross-group pairs.
    The direction of the resulting bias depends on the correlation structure; characterizing its magnitude for specific ranking systems is an open question.
\end{enumerate}

\subsection{Future Work}
\label{sec:disc-future}

\begin{enumerate}
  \item \textbf{BISG refinement.}
    The current BISG model uses generic U.S.\ Census frequency tables without adaptation to the target member population.
    Calibrating the model for this population would improve estimation accuracy.
    Methods include fine-tuning the prior using Self-ID data (with appropriate privacy protections) and incorporating additional demographic signals beyond surname and geolocation.

  \item \textbf{Uncertainty quantification.}
    Current measurements report bootstrap confidence intervals for the final disparity statistics, but these intervals capture only sampling variability.
    They do not account for uncertainty in the BISG estimates themselves.
    Propagating BISG prediction uncertainty through the estimators would provide a more complete picture of measurement reliability, but this is technically challenging because it requires characterizing the joint distribution of BISG errors across the test population.

  \item \textbf{Cross-jurisdictional extension.}
    Extending PPRE to non-U.S.\ jurisdictions depends on the availability of suitable demographic data.
    The EU's fragmented equality data landscape presents particular challenges, and any extension must also navigate differing legal frameworks governing the use of demographic data for algorithmic fairness evaluation.
    The signal-processing pillar of PPRE is modular enough to accommodate alternative demographic models, but the practical barriers are substantial.
\end{enumerate}


\section{Conclusion}
\label{sec:conc}

We presented PPRE: a production-deployed system for privacy-preserving race/ethnicity fairness measurement. PPRE monitors AI systems for equal treatment when demographic data is severely limited and privacy constraints are strict.
The system rests on three pillars.

\emph{Demographic signal processing.} BISG generates probabilistic race/ethnicity posteriors from surname and geolocation; sparse Self-ID records provide calibration.
Probabilistic disparity estimators replace hard demographic indicators with soft ones, generalizing standard fairness metrics to operate on probability distributions.
This gives weighted-average statistics that are both accurate and compatible with encrypted computation.

\emph{Privacy primitives and extensions.} Commutative encryption enables private set intersection. Additive homomorphic encryption supports computation on ciphertexts. Local differential privacy protects Self-ID records at the input stage.
A catalog of five HE adaptations extends the toolkit to support the division, comparison, and hypothesis testing that production fairness metrics require.
These adaptations are general-purpose: they apply to any ratio-form statistic under additive HE, not only to fairness metrics or race/ethnicity measurement.

\emph{Protocol composition.} A custom two-party computation protocol composes these elements so that neither party ever observes the other's plaintext inputs, with all member-level data ephemeral throughout.
Production fairness measurements of two AI ranking systems using LOT and MQOS-NDCG found no statistically significant race-related biases.
All measurements ran entirely in ciphertext space with bootstrap significance testing.

PPRE serves as an existence proof that privatized multi-party fairness measurement is feasible at industrial scale.
The implementation experience also reveals where the hard problems lie: testing and debugging multiparty flows incurs costs at least an order of magnitude higher than single-party equivalents.
And geographically scoped demographic models introduce population-selection effects that must be carefully characterized.
The design principles, protocol architecture, and engineering patterns documented here offer a concrete reference for institutions building similar infrastructure.

\subsection*{Acknowledgements}
This work was the result of a fruitful and massive collaboration between the responsible AI, data privacy, and legal teams.
We would like to thank Rahul Tandra, Joaquin Qui\~nonero Candela, and Ya Xu for their leadership support; Daniel Tweed-Kent, Matthew Baird, Sam Gong, and Igor Perisic for their feedback; and Ray Ortigas, Helo\"ise Logan, Sara Harrington, Jen Carmenate, and Jon Adams for their contributions and support.
Finally, we express our deep gratitude to those members who trusted us with their Self-ID data.
This work would not have been possible without their contribution.

\bibliographystyle{acm-styles/ACM-Reference-Format}
\bibliography{refs}

\appendix

\section{Cryptographic and Statistical Primitives}
\label{sec:appendix-primitives}

This appendix provides complete formal definitions of the cryptographic and statistical primitives summarized in \sectionref{pets}.

\subsection{Commutative Encryption}

A commutative encryption scheme $\com = (\comkeygen, \comenc, \comdec)$ is a symmetric encryption scheme (with a deterministic encryption algorithm) over message space $\M$:
\begin{itemize}
\item $\sk \gets \comkeygen(1^\secr)$
\item $\ct \gets \comenc_\sk(m)$
\item $m / \bot \gets \comdec_\sk( \ct)$
\item For any two keys $\sk_1, \sk_2$ and any message $m \in \M$,
	$$\comenc_{\sk_1}(\comenc_{\sk_2}(m)) = \comenc_{\sk_2}(\comenc_{\sk_1}(m))$$
\end{itemize}

A popular instantiation is the Pohlig-Hellman cipher over an Elliptic Curve group $\G$ in which the Decisional Diffie-Hellman (DDH) assumption holds. A random oracle $\RO$ maps arbitrary messages into $\G$. \sectionref{impl} details our specific instantiation.

This commutativity property, combined with a fresh $\RO$ instance per protocol execution, serves as the building block for private set intersection (PSI) protocols~\cite{Meadows86,HFH99}. PSI lets two parties learn only the intersection of their sets without exposing non-shared elements. \sectionref{ppre-system} applies this methodology to join the two parties' datasets on member identifiers.

\subsection{Additive Homomorphic Encryption}

An additive homomorphic encryption scheme $\ahe=(\ahekeygen,\aheenc,\aheadd,\ahedec)$ over message space $\M$:
\begin{itemize}
\item $(\pk, \sk) \gets \ahekeygen(1^\secr)$
\item $\ct \gets \aheenc_\pk(m;r)$
\item $m / \bot \gets \ahedec_\sk( \ct)$
\item Homomorphic addition: $\ahedec_\sk(\aheadd(\aheenc_\pk(m_1), \aheenc_\pk(m_2))) = (m_1+m_2)$  $\forall m_1, m_2\in \M$.
\item Homomorphic multiplication with constant: $\ahedec_\sk(\aheadd(c, \aheenc_\pk(m))) = (c\cdot m)$  $\forall c, m \in \M$.
\end{itemize}
We overload $\aheadd$ to denote both homomorphic addition and scalar multiplication. Each homomorphic evaluation is implicitly followed by a refresh operation: adding an encryption of zero so the resulting ciphertext's randomness is independent of the originals. For Paillier encryption~\cite{Paillier99}, which rests on the Decisional Composite Residuosity assumption, homomorphically evaluated ciphertexts are statistically identical to fresh ones. See~\cite{Paillier99} for formal correctness and CPA security definitions.

We also use traditional randomized symmetric encryption $(\symkeygen, \symenc, \symdec)$.

\subsection{Differential Privacy: Randomized Response Mechanism}

We use the traditional local DP algorithm called randomized response~\cite{Warner65}. For records taking one of $k$ values $\{1,\ldots,k\}$, the mechanism $\mathcal{M}: \{1,\ldots,k\} \rightarrow \{1,\ldots,k\}$ retains input $i$ with probability $\frac{e^\epsilon}{e^\epsilon + k - 1}$ and flips it to each of the other $(k-1)$ values with probability $\frac{1}{e^\epsilon + k - 1}$:
\begin{align*}
\mathsf{Pr}[\mathcal{M}(i)  = i] &= \frac{e^\epsilon}{e^\epsilon + k - 1}
\\
\mathsf{Pr}[\mathcal{M}(i)  \in \{ 1, \cdots k\} \setminus \{i\} ] &= \frac{k-1}{e^\epsilon + k - 1} .
\end{align*}
The privacy parameter $\epsilon$ quantifies the tradeoff: a larger $\epsilon$ means a higher probability that any record was not falsified.
\section{Additional Metric Algorithms}
\label{sec:appendix-metrics}

This appendix specifies two additional fairness metrics whose PPRE adaptations have been fully designed but are not yet deployed in production.
Both follow the same three-part structure as the deployed metrics in \sectionref{metrics}: standard definition, probabilistic adaptation, and HE computation strategy.

\subsection{POT: Pointwise Outcome Test}
\label{sec:pot}

POT is a candidate-side metric for evaluating score-producing models.
LOT evaluates ranking fairness for listing models via adjacent-rank comparisons. POT evaluates \emph{calibration fairness} for scoring models: it checks whether the expected outcome at a given score level is the same across demographic groups.

\paragraph{Standard definition.}
Given observed triplets $(S_i, Y_i, R_i)$ (predicted score, realized outcome, and group label), POT tests whether the regression function $\mathbb{E}[Y \mid S = s, R = j]$ is approximately equal across groups $j$.
The test uses nonparametric (Nadaraya-Watson) kernel regression at $Q$ reference points $\{s_1, \ldots, s_Q\}$ (typically score deciles):
\begin{equation}
    \hat{m}_j(s_q) = \frac{\sum_{i=1}^{n} \mathbb{I}(R_i = j) \cdot K_h(S_i - s_q) \cdot Y_i}{\sum_{i=1}^{n} \mathbb{I}(R_i = j) \cdot K_h(S_i - s_q)}
    \label{eq:pot}
\end{equation}
where $K_h(\cdot)$ is a Gaussian kernel with bandwidth $h$.
Disparities across groups are summarized by a Discounted Total Impact (DTI) metric:
\begin{equation}
    \text{DTI}_{a,b} = \sum_{q=1}^{Q} w_q \cdot \bigl(\hat{m}_a(s_q) - \hat{m}_b(s_q)\bigr)
    \label{eq:dti}
\end{equation}
where $w_q$ are position-dependent discount weights.

\paragraph{Probabilistic adaptation.}
Replace the hard group indicator $\mathbb{I}(R_i = j)$ with the BISG posterior $\mathsf{Pr}(R_i = j)$. This modulates each individual's contribution to the group-specific regression by their probability of membership:
\begin{equation}
    \hat{m}_j(s_q) = \frac{\sum_{i=1}^{n} \mathsf{Pr}(R_i = j) \cdot K_h(S_i - s_q) \cdot Y_i}{\sum_{i=1}^{n} \mathsf{Pr}(R_i = j) \cdot K_h(S_i - s_q)}.
    \label{eq:pot-prob}
\end{equation}
This is an instance of the model performance disparity estimator (Equation~\ref{eq:soft-ai-sample-mean}) with the test value modulated by the kernel weight.

\paragraph{HE computation strategy.}
POT composes three adaptation patterns:
\begin{enumerate}
    \item \textbf{Pre-encryption comparison} (\sectionref{pre-encryption-comparison}). P2 holds the scores $S_i$, outcomes $Y_i$, and the kernel function $K_h$.
    P2 pre-computes the kernel-weighted outcome products $K_h(S_i - s_q) \cdot Y_i$ and kernel weights $K_h(S_i - s_q)$ at each reference point $s_q$ in plaintext, then encrypts both quantities.

    \item \textbf{Ratio estimation} (\sectionref{ratio-estimation}). After the encrypted join, P1 computes per-group, per-reference-point aggregates:
    \begin{align*}
        \aheenc(\text{Num}_{j,q}) &= \bigoplus_{i=1}^{n} \mathsf{Pr}(R_i = j) \odot \aheenc\!\bigl(K_h(S_i - s_q) \cdot Y_i\bigr) \\
        \aheenc(\text{Den}_{j,q}) &= \bigoplus_{i=1}^{n} \mathsf{Pr}(R_i = j) \odot \aheenc\!\bigl(K_h(S_i - s_q)\bigr).
    \end{align*}
    Random masking is applied before transfer to P2.

    \item \textbf{Plaintext post-processing.} P2 decrypts, divides to obtain $\hat{m}_j(s_q)$ at each reference point, computes the DTI in plaintext, and applies bootstrap significance testing (\sectionref{bootstrap}) to assess statistical significance.
\end{enumerate}

LOT and POT share a key structural property: demographic weights come entirely from P1's data, and outcome values come entirely from P2's data. For LOT these outcome values are score differences; for POT they are kernel-weighted outcomes. This clean partition maps directly onto the HE computation model: P1 multiplies plaintext weights against P2's ciphertexts and sums.

\subsection{MQOS-AUC: Minimum Quality of Service-AUC}
\label{sec:mqos-auc}

MQOS-AUC is a viewer-side metric that compares group-specific Area Under the ROC Curve (AUC) against the global AUC.
AUC measures the probability that a model ranks a randomly chosen positive instance above a randomly chosen negative instance. MQOS-AUC checks whether this discriminative quality differs across demographic groups.

\paragraph{Fairness criterion.}
MQOS-AUC flags a disparity when:
\begin{equation}
    \text{AUC}_{\text{global}} - \text{AUC}[j] > \tau, \quad \text{for any group } j
    \label{eq:mqos-auc-criterion}
\end{equation}
where $\tau$ is a pre-specified tolerance threshold.
A violation indicates that group $j$ receives meaningfully worse model discrimination quality than the overall population.

PPRE implements two algorithm variants for computing probabilistic group-level AUC under HE constraints.
The choice between them depends on dataset size and computational budget.

\subsubsection{Algorithm 1: Probabilistic ROC-AUC}
\label{sec:auc-roc}

This algorithm constructs the group-specific ROC curve via a single sorted pass through the data and computes AUC using the trapezoidal rule.

\paragraph{Probabilistic TPR and FPR.}
For group $j$, the probabilistic true positive rate and false positive rate at threshold $t$ are:
\begin{align}
    \text{TPR}_j(t) &= \frac{\sum_{i:\, Y_i = 1} \mathsf{Pr}(R_i = j) \cdot \mathbb{I}(s_i \geq t)}{\sum_{i:\, Y_i = 1} \mathsf{Pr}(R_i = j)} \label{eq:prob-tpr} \\[4pt]
    \text{FPR}_j(t) &= \frac{\sum_{i:\, Y_i = 0} \mathsf{Pr}(R_i = j) \cdot \mathbb{I}(s_i \geq t)}{\sum_{i:\, Y_i = 0} \mathsf{Pr}(R_i = j)} \label{eq:prob-fpr}
\end{align}
where $s_i$ is the model score for individual $i$ and $Y_i$ is the true label.

\paragraph{AUC via trapezoidal rule.}
Using all $n$ distinct score values as thresholds, sorted in descending order, the probabilistic AUC for group $j$ is:
\begin{equation}
    \text{AUC}[j] = \sum_{m=1}^{n} \tfrac{1}{2}\bigl(\text{FPR}_j(t_{m+1}) - \text{FPR}_j(t_m)\bigr) \cdot \bigl(\text{TPR}_j(t_m) + \text{TPR}_j(t_{m+1})\bigr).
    \label{eq:auc-trapezoid}
\end{equation}

\paragraph{HE computation strategy.}
The algorithm requires sorting instances by score and incrementally computing cumulative weighted sums.
Sorting requires comparison, which ciphertext space does not support.
P2 holds the model scores and labels, so P2 sorts unilaterally.

\begin{enumerate}
    \item \textbf{Pre-encryption comparison} (\sectionref{pre-encryption-comparison}). P2 sorts all instances in descending order by score and pre-computes label-based indicators.
    For each instance $i$ in sorted order, P2 encrypts a tuple containing the label indicator and a constant~$1$.

    \item \textbf{Cumulative weighted aggregation.} P1 processes instances in the pre-sorted order.
    For each group $j$, P1 maintains running weighted sums of true positives and false positives using homomorphic scalar multiplication (by $\mathsf{Pr}(R_i = j)$) and addition.
    At each step, the cumulative sums correspond to $\text{TPR}_j(t)$ and $\text{FPR}_j(t)$ at the current threshold, up to normalization.

    \item \textbf{Trapezoidal area increments.} Each step's contribution to the AUC integral (Equation~\ref{eq:auc-trapezoid}) is a linear combination of the current and previous cumulative sums, computable in HE.
    The denominators (total positive and negative weights per group) are accumulated in parallel via ratio estimation (\sectionref{ratio-estimation}).

    \item \textbf{Decryption and normalization.} P2 decrypts the accumulated numerator and denominator terms and computes the final AUC values in plaintext.
\end{enumerate}

The computational complexity is $O(n \log n)$ (dominated by sorting) with $O(k)$ additional memory where $k$ is the number of demographic groups.

\subsubsection{Algorithm 2: Sampled Mann-Whitney AUC}
\label{sec:auc-mw}

This algorithm uses the Mann-Whitney U statistic. The statistic provides an equivalent definition of AUC: the probability that a randomly chosen positive instance is scored higher than a randomly chosen negative instance.

\paragraph{Probabilistic Mann-Whitney AUC.}
For a set $S$ of randomly sampled positive-negative pairs, the probabilistic group-$j$ AUC is:
\begin{equation}
    \text{AUC}[j] = \frac{\sum_{(i,\ell) \in S} \mathsf{Pr}(R_i = j) \cdot \mathsf{Pr}(R_\ell = j) \cdot \mathbb{I}(s_i > s_\ell)}{\sum_{(i,\ell) \in S} \mathsf{Pr}(R_i = j) \cdot \mathsf{Pr}(R_\ell = j)}
    \label{eq:auc-mw}
\end{equation}
where each pair $(i, \ell) \in S$ satisfies $Y_i = 1$ (positive) and $Y_\ell = 0$ (negative).
Group membership probabilities determine each pair's contribution to group $j$'s AUC.

\paragraph{HE computation strategy.}
The sampled Mann-Whitney algorithm requires comparison indicators and pair-weight products. Neither is directly computable in HE. The adaptation exploits the data partition:

\begin{enumerate}
    \item \textbf{Pre-encryption comparison} (\sectionref{pre-encryption-comparison}). P2 generates random positive-negative pairs from its data, pre-computes the comparison indicator $\mathbb{I}(s_i > s_\ell)$ in plaintext, and encrypts the result.

    \item \textbf{Weight computation.} P1 holds the race probabilities for both members of each pair.
    The pair weight $\mathsf{Pr}(R_i = j) \cdot \mathsf{Pr}(R_\ell = j)$ is a product of two plaintext values held by P1, computed entirely in plaintext.

    \item \textbf{Ratio estimation} (\sectionref{ratio-estimation}). P1 computes the weighted ciphertext sum:
    \begin{align*}
        \aheenc(\text{Num}[j]) &= \bigoplus_{(i,\ell) \in S} \bigl[\mathsf{Pr}(R_i = j) \cdot \mathsf{Pr}(R_\ell = j)\bigr] \odot \aheenc\!\bigl(\mathbb{I}(s_i > s_\ell)\bigr)
    \end{align*}
    with the denominator accumulated in parallel.
    Random masking is applied before transfer; P2 decrypts and divides.
\end{enumerate}

The computational complexity is $O(m)$, where $m = |S|$ is the number of sampled pairs. This cost is independent of the total dataset size $n$. The sampled variant is therefore preferable for very large datasets where $O(n \log n)$ or $O(n^2)$ becomes expensive.

\subsubsection{Algorithm Comparison}
\label{sec:auc-comparison}

\begin{table}[h]
\centering
\caption{Comparison of AUC algorithm variants for PPRE.}
\label{tab:auc-comparison}
\begin{tabular}{lcc}
\toprule
\textbf{Property} & \textbf{Probabilistic ROC-AUC} & \textbf{Sampled Mann-Whitney} \\
\midrule
Complexity & $O(n \log n)$ & $O(m)$ \\
Memory & $O(k)$ & $O(m)$ \\
Exactness & Exact (numerical precision) & Approximate (sampling error) \\
Best suited for & General purpose & Very large datasets \\
\bottomrule
\end{tabular}
\end{table}

Both algorithms use the same HE adaptation strategy: P2 pre-computes all comparison and sorting operations, and P1 performs weighted linear aggregation in ciphertext space. Dataset scale and acceptable approximation error determine the choice between them.
\section{Additional Disparity Estimator Types}
\label{sec:appendix-estimators}

The model performance disparity estimator (Equation~\ref{eq:soft-ai-sample-mean}) covers the three fairness metrics deployed in production.
Two additional estimator types extend the framework to other measurement tasks.

\subsection{Output Metric Disparity Estimator}

This estimator applies to \emph{outcome equity measurements} or \emph{product equity measurements}: disaggregated evaluations of a scalar outcome $Y$ that is not derived from a model's predictions.
It computes group-wise sample means on $Y$ directly:
\begin{equation}
    \mu[j] = \frac{\sum_{i=1}^{n} \mathsf{Pr}(R_i=j) \cdot Y_i}{\sum_{i=1}^{n} \mathsf{Pr}(R_i=j)}, \; \forall j \in \{1, \ldots, k\}
    \label{eq:soft-sample-mean}
\end{equation}
This is a special case of Equation~\ref{eq:soft-ai-sample-mean} with $f(Y_i, \hat{Y}_i) = Y_i$.
It is linear in the test values and computable under additive HE via the same ratio estimation pattern.

\subsection{Probabilistic Count Disparity Estimator}

This is an \emph{inclusiveness measurement}: it estimates whether at least a threshold number $T$ of members from a race/ethnicity group are present in a member pool.
For threshold $T=1$ and group $j$:
\begin{align}
    \mathsf{P}_{Equity} &= \mathsf{Pr}(N_{j} \geq 1 | N_{pool}) = 1 - \mathsf{Pr}(N_{j} = 0 | N_{pool}) \notag \\
    \mathsf{P}_{Equity} &= 1 - \prod_i^{N_{pool}}\Big(1 - \mathsf{Pr}(R_i = j)\Big) \;.
    \label{eq:soft-count}
\end{align}
The measuring party compares this to a specified probability level, e.g. $\mathsf{P}_{Equity} \gtrless 0.90$, to assess with $90\%$ certainty whether the pool includes at least one member from the group of interest.
When $T>1$, $N_{j} \sim \text{PoiBin}$, requiring more detailed calculations.

This estimator is non-linear, so it cannot be computed under additive HE.
However, it depends only on race/ethnicity probabilities known to $P_1$ and pool membership from $P_2$, making it directly feasible under the two-party protocol without homomorphic computation.
\section{BISG Quality Validation}
\label{sec:bisg-validation}

We validated BISG outputs by comparing them against Self-ID survey responses using cross-entropy.
The Self-ID survey covers a self-selected subpopulation, making this an imperfect comparison, but it provides the best available ground truth.

We compared BISG variants using different input combinations:
\begin{itemize}
    \item \textbf{Baseline BISG}: surname and geolocation (ZIP code) only.
    \item \textbf{BIFSG}: BISG extended with first name distributions from the Harvard Dataverse survey~\cite{tzioumis_2018}.
    \item \textbf{BISF}: surname and first name only, without geolocation.
\end{itemize}

\begin{figure}[ht!]
    \centering
    \includegraphics[width=0.65\linewidth]{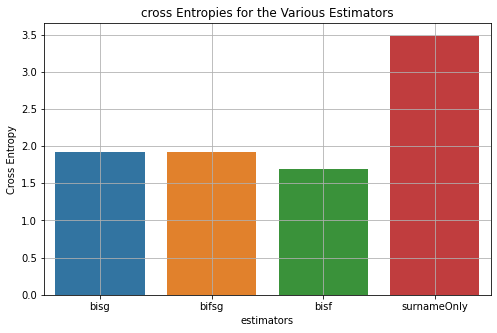}
    \caption{Comparing the cross-entropy of BISG variants against Self-ID survey responses on a member sub-sample. Lower is better.}
    \label{fig:bisg-xent-cmp}
\end{figure}

Figure~\ref{fig:bisg-xent-cmp} shows the results.
BIFSG does not meaningfully improve over baseline BISG; the first name signal adds minimal discriminative power.
BISF yields slightly lower overall cross-entropy but diminished performance for smaller groups with lower base rates. Geolocation provides critical signal for geographically concentrated groups, and removing it hurts those estimates.
We therefore use baseline BISG for all production measurements.

\section{Sample Disparity Measurements}
\label{sec:sample-measurements}

We present sample measurements that demonstrate end-to-end PPRE operation using the basic disparity estimators from \sectionref{audit-func}.
These measurements use the output metric and model performance estimators on illustrative tasks.

\subsection{Seniority Share (Output Metric)}

Using the output metric disparity estimator (Equation~\ref{eq:soft-sample-mean}), we estimated the fraction of members holding senior roles across all six OMB race/ethnicity categories.
Here $Y_i$ is a seniority indicator computed from member profile entries.
Table~\ref{tab:egri-bisg} shows the results for a random sample of the U.S.\ member population.

\begin{table}[ht!]
    \centering
    \begin{tabular}{lc}
    \toprule
    \textbf{Race/Ethnicity Group} & \textbf{Estimated Seniority Share} \\
    \midrule
    White & 0.2334 \\
    Black/African American & 0.2191 \\
    Hispanic/Latino & 0.2107 \\
    Asian & 0.2285 \\
    American Indian/Alaska Native & 0.2243 \\
    Native Hawaiian/Pacific Islander & 0.2198 \\
    \bottomrule
    \end{tabular}
    \caption{PPRE seniority share measurement across all six OMB race/ethnicity groups for a random sample of the U.S.\ member population.}
    \label{tab:egri-bisg}
\end{table}

The White and Black group estimates agree in relative order with a separate internal analysis~\cite{Baird2024} that used Self-ID survey responses only ($W\!:\!0.3459 > B/A\!:\!0.3337$).
The estimated magnitudes differ because the two measurements cover statistically dissimilar populations: PPRE covers the broader U.S.\ member population, while the comparison measurement is restricted to Self-ID respondents.

\subsection{False Positive Rate (Model Performance)}

Using the model performance disparity estimator (Equation~\ref{eq:soft-ai-sample-mean}), we computed group-wise false positive rates for a hypothetical AI classification model.
Here $Y$ is the ground truth classification and $\hat{Y}$ is the model's predicted classification.

\begin{table}[ht!]
    \centering
    \begin{tabular}{lc}
        \toprule
        \textbf{Race/Ethnicity Group} & \textbf{Estimated FPR} \\
        \midrule
        White & 0.00121 \\
        Black/African American & 0.00126 \\
        \bottomrule
    \end{tabular}
    \caption{PPRE group-wise false positive rates for a hypothetical classification model.}
    \label{tab:ero-bisg}
\end{table}

\end{document}